\documentclass[12pt, a4paper]{article}
\usepackage{amsmath,amsfonts,amssymb}
\usepackage{graphicx}
\usepackage{setspace}
\usepackage{tocloft}
\usepackage{multirow,multicol}
\usepackage{array}

\title{Role of Class-specific Features in Various Classification Frameworks for Human Epithelial (HEp-2) Cell Images}

\author{Vibha~Gupta, Arnav~Bhavsar\\School of Computer and Electrical Engineering\\
	Indian Institute of Technology Mandi, India}
	
	\date{}

\begin{document} 
\maketitle

\begin{abstract}
The antinuclear antibody detection with human epithelial cells is a popular approach for autoimmune diseases diagnosis. The manual evaluation demands time, effort and capital, automation in screening can greatly aid the physicians in these respects. In this work, we employ simple, efficient and visually more interpretable, class-specific features which defined based on the visual characteristics of each class. We believe that defining features with a good visual interpretation, is indeed important in a scenario, where such an approach is used in an interactive CAD system for pathologists. Considering that problem consists of few classes, and our rather simplistic feature definitions, frameworks can be structured as hierarchies of various binary classifiers. These variants include frameworks which are earlier explored and some which are not explored for this task. We perform various experiments which include traditional texture features and demonstrate the effectiveness of class-specific features in various frameworks. We make insightful comparisons between different types of classification frameworks given their silent aspects and pros and cons over each other. We also demonstrate an experiment with only intermediates samples for testing. The proposed work yields encouraging results with respect to the state-of-the-art and highlights the role of class-specific features in different classification frameworks.


\end{abstract}
{{\it keywords:} Antinuclear antibody (ANA) detection; Class-specific features; Classification frameworks; Support vector machine (SVM); Two-stage frameworks; Computer-aided diagnosis (CAD)}


\begin{spacing}{2}   

\section{Introduction}
\label{sect:intro}  
Autoimmune disorders are immunological conditions wherein the immune system erroneously attack the healthy self-cells of the tissues as the compromised immune cells consider these self-cells foreign or dangerous. In this condition, the body releases various antibodies, one of them being Antinuclear antibodies (ANA). These immunoglobulins recognize autologous components (simple proteins or conjugated proteins) within the nuclear and cytosolic compartment of the cell as antigenic~\cite{Cabiedes}. The specificity of antinuclear antibodies type differs based on the stimulus driven in respective kind of autoimmune disorder. Thus it becomes a great concern; to identify presence of these ANA, so that the medical supervision could enable the patient at the earliest to reduce the extent of destruction within their body. The presence of ANAs in patient's blood is confirmed by screening tests. Although there are many tests for the detection of ANAs, the most common tests used for screening include indirect immunofluorescence (IIF) and enzyme-linked immunosorbent assay (ELISA). Due to better specificity and sensitivity of IIF imaging, it is prominently used in diagnosis of specific autoimmune diseases by evaluating its corresponding antibodies within the patient serum. 

The presence of auto-antibodies can be easily assessed, as the antibodies bind with the nuclear proteinaceous components of HEp-2 cells, considering them as antigens (Ag). This Ag-Ab (Antigen - Antibody) complex formation can be visualized through Indirect Immunofluorescence microscopy and thus the test can also be greatly qualitative and easily decisive. 
\begin{figure}[h!]
	\centering
	
	\resizebox{13.03cm}{!}{
		\includegraphics[width=150pt, height=135pt]{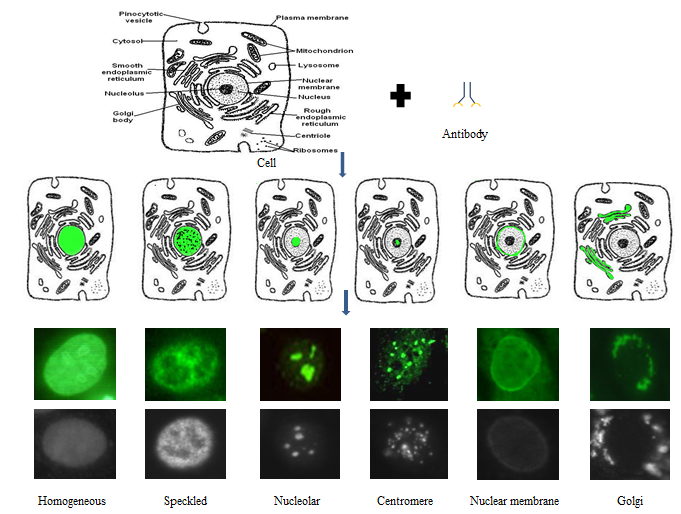}}
	\caption{\label{fig:one}Fluorescence patterns at distinct locations.}
\end{figure}
The molecular complex manifests different nuclear staining patterns of fluorescence in HEp-2 cells. The reaction between certain antigenic substances like (for e.g. proteins, phospholipids etc.) specific to a particular type of cell organelle and the corresponding antibody (Ab) from the patient serum is demonstrated in Figure ~\ref{fig:one}, wherein the antibody detected in various organelles of the cell is visualized by a specific fluorescent pattern at corresponding sites. 
The IIF technique generates multitude of patterns of stained cells, post incubating patient's serum with HEp-2 cell line and fluorescence labeling of the anti-human immunoglobulins~\cite{sack}. This technique produces different staining patterns in cells depending on presence of different antibodies. Figure~\ref{fig:ten} describes the complete procedure of Indirect Immunofluorescence Imaging using HEp-2 cells.

Diagnostic implications based ANA IIF images involves mitotic cells recognition along with classification of the interphase staining patterns. This recognition step seems significant for various aspects. Firstly, it portrays a clean and well prepared sample~\cite{bradwell}. More on importance of mitotic detection are detailed in references~\cite{foggia19,iannello}. While the importance of 
mitotic detection (a binary classification problem) is shown in some recent works~\cite{foggia19,iannello}, here, in this work, we mainly focus on the multi-class interphase cell classification problems. This problem itself important from the perspective of Auto-immune disease diagnosis. Indeed, the usefulness of such interphase classification  is indicated by the presence of relatively large dataset that we use in this work~\cite{hobson2015}, which only contains cell images of interphase patterns.

It is observed that only a hand-full of staining patterns within this IIF images are clinically significant ~\cite{Kumar}. Some of these are Homogeneous, Speckled, Nucleolar, Centromere, Nuclear Membrane and Golgi that are most often detected by IIF on HEp-2 cells in the sera of patients with autoimmune disease. The manual test evaluation of HEp-2 cells in laboratory demands expertise and capital. Moreover, the opinions of medical experts are often subjective and not easily quantifiable. Due to this dependency on the subjectivity of the specialist, the manual visual analysis of HEp-2 cell patterns suffers from lower reproducibility and reliability of obtained result. This further translates the manual test evaluation into economically taxing healthcare diagnostic system. The existing literature also steadily highlights gross inter and intra-laboratory variability ~\cite{makadia}.
Development of computer-aided diagnosis (CAD) system can aid the physician in multiple ways: These can (i) augment the capability of physician, thus reduce the error by acting as second reader, (ii) facilitate mass screening campaigns by focusing the attention on the most relevant cases, and (iii) provide an effective and simple tool for instruction and training of specialized medical personnel. This highlights a need for reliable and efficient algorithms for IIF cell image classification. 

A vital step of ANA HEp-2 test includes interpretation of obtained stained patterns of HEp-2 cells for establishing a correct diagnosis. This test produces diverse staining patterns due to staining of different cellular organelles like nucleoli, nucleus, cytoplasm or chromosomes which can show disparity in size, shape, number. This allows interpreters to distinguish among staining patterns characteristic for different autoimmune diseases.

To automate the IIF diagnostic procedure, various machine learning methods have been proposed in context of ANA testing. Such approaches can be decomposed into feature extraction and classification components. In the existing approaches, for automatic classification, commonly available features for describing morphology and texture were extracted~\cite{Ghosh,Agrawal,Piro}. However, considering that there is an ample expert knowledge regarding the morphological properties of stained cell region (such as number, size, localization and shape of certain physical traits), we believe that this may benefit the feature extraction in classification task and help to characterize the patterns in a better manner~\cite{Perner}. Moreover, we suggest that such features can be more simplistically defined and can be more interpretable than the abstract features which are typically borrowed from the computer vision community.

Furthermore, it is interesting to ask the question that with the same features, how does the choice of classification structure matter. In this regard, while existing research in HEp-2 cell image classification has considered a various classifiers, there is no work with a targeted comparisons between such classifiers. Moreover, some useful ensemble classification frameworks, have not been explored for this task. In this work, we also explore a review and applications of a variety of such frameworks. We elaborate on the scope of this work below.

\subsection{Scope of the work}
In this work, we propose the idea of extracting relatively small set of class-specific features separately for each class. The main focus of our feature extraction approach is to utilize morphological properties of stained-cell regions. We extract various features, that capture the patterns that are apparent through visual traits. In other words, we utilize the existing information on visual characteristics of each class, that is further used to extract class specific features. Our feature definitions are quite simplistic, and visually more interpretable. We believe that defining features with a good visual interpretation, is indeed important from the point of view of manual interpretation of features (in a scenario, where such an approach is used in an interactive CAD system for pathologists). 
To show the effectiveness of the proposed features, we also compare the classification performance using these with some traditional texture features and some state-of-the-art approaches.  
\begin{figure}[t!]
	\centering
	\resizebox{14.03cm}{!}{
		\includegraphics[width=160pt, height=35pt]{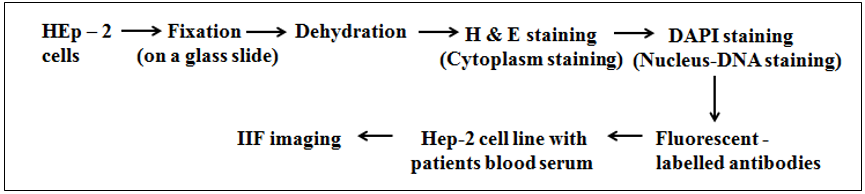}}
	\caption{\label{fig:ten}Procedure of Indirect Immunofluorescence Imaging using HEP-2 cells.}
\end{figure}

Considering that the problem consists of a small number of classes (which is, arguably, true for many medical image classification tasks), and considering our rather simplistic feature definitions, we suggest that the classification framework can be designed in many ways e.g. structures where categories are arranged as a collection or hierarchies of various binary classifiers. This can also be interpreted as forms of ensemble classification. We compare a variety of such structures, and demonstrate their silent aspects, and pros and cons over each other. This includes variants of classification frameworks which are earlier explored (for HEP-2 problem), and some frameworks which are not explored for this task. For example, frameworks such as random uniform forest~\cite{ciss}, adaboost~\cite{freund19}, and some ensemble structure of support vector machines (SVM), which we consider here, have not been explored for HEp-2 cell problem. 
Thus, such a review demonstration of a variety of classifiers also contributes to the progress in this area. 

Interestingly, we demonstrate that while the performance of the classification strategies is different with the class-specific features, the superiority of class-specific features is maintained across all classification strategies.  Hence, our research adds to a relatively limited but growing body of work research on HEp-2 cell classification exploring classification frameworks utilizing class-specific features from HEp-2 cell image classification point of view.

Furthermore, the proposed methods are evaluated using ICPR-2014 (13596 cell images)~\cite{Hobson} data which consists more images than some earlier datasets (e.g. ICPR-2012 (1455 cell images)~\cite{Saggese}).

\subsection{Related work}
This section discusses the previous work that has been done in past few years, including state-of-the-art methods. All the discussed methods aimed to automate the IIF diagnostic procedure for ANA testing. Perner et. al.~\cite{Perner} performed an early attempt in this context. They extracted various features from cell  region using multilevel gray thresholding, and used data mining algorithm to come cross relevant feature set among large feature sets. In~\cite{Huang}, fourteen textural and statistical features and self-organizing map (SOM) were utilized to classify the fluorescence patterns. Hsieh et. al.~\cite{Hsieh} utilized learning vector quantization (LVQ) with eight textural features  to identify the fluorescence pattern. All the above discussed methods were evaluated on private dataset which makes performance comparison  very tough. The contests which were recently held in conjunction with the ICPR-2012, ICIP-2013, and ICPR-2014 on HEp-2 cell classification ~\cite{Saggese,Foggia,hobson20,hobson201} released public datasets which are suitable for evaluation of methods as a part of relevant contests. These contests have given a strong stimulus for the development of automatic algorithms for the task of HEp-2 cell image classification. In this context, several new approaches have been proposed for cell classification and evaluated on common basis. In addition to these HEp-2 cell datasets, Wiliem et al.~\cite{wiliem20} released SNP HEp-2 cell dataset (publicly available) as benchmarking platform. It contains 1884 cell images of five patterns: homogeneous, nucleolar, centromere, coarse speckled, fine speckled. The dataset was divided into training and testing sets, which contain 905 and 979 images, respectively. 

We note however, that the datasets with ICPR-2012, SNP HEp-2 were of a much smaller size (1455, 1884 images), and the larger dataset viz. one with ICPR-2014 containing 13659 images, is generally used by the state-of-the-art methods. We too use the same in this work. Table~\ref{table15} provides the details of datasets available for HEp-2 cell problem. More recent reviews of this topic were provided by~\cite{hobson2015,hobson2016}. The reader can refer to~\cite{hobson1} for an overview of the literature and of the open challenges in this research area. Below, we discuss some specific works in terms of features and classifiers. 

\begin{table*}[h!]
	\centering
	\caption{Existing HEp-2 cell datasets.}
	\label{table15} 
	\resizebox{13.80cm}{!}{
		\begin{tabular}{|c|c|c|c|c|c|c|}
			\hline
			\multirow{2}{*}{\textbf{}} & \multicolumn{2}{c|}{\textbf{\begin{tabular}[c]{@{}c@{}}Number of Images\\ in ICPR-2012\end{tabular}}} & \multicolumn{2}{c|}{\textbf{\begin{tabular}[c]{@{}c@{}}Number of Images\\ in SNHEp-2\end{tabular}}} & \multicolumn{2}{c|}{\textbf{\begin{tabular}[c]{@{}c@{}}Number of Images\\ in ICPR-2014/ICIP-2013\end{tabular}}} \\ \cline{2-7} 
			& \multicolumn{2}{c|}{Using 28 specimen images}                                                         & \multicolumn{2}{c|}{Using 40 specimen images}                                                       & \multicolumn{2}{c|}{Using 83 specimen images}                                                                   \\ \hline
			\textbf{Patterns}          & \textbf{Train}                                     & \textbf{Test}                                    & \textbf{Train}                                    & \textbf{Test}                                   & \textbf{Train}                                          & \textbf{Test}                                         \\ \hline
			\textbf{Homogeneous}       & 150                                                & 180                                              & 172                                               & 188                                             & 2494                                                    & -                                                     \\ \hline
			\textbf{Speckled}          & -                                                  & -                                                & -                                                 & -                                               & 2831                                                    & -                                                     \\ \hline
			\textbf{Coarse speckled}   & 109                                                & 101                                              & 166                                               & 187                                             & -                                                       & -                                                     \\ \hline
			\textbf{Fine speckled}     & 94                                                 & 114                                              & 188                                               & 191                                             & -                                                       & -                                                     \\ \hline
			\textbf{Nucleolar}         & 102                                                & 139                                              & 194                                               & 139                                             & 2598                                                    & -                                                     \\ \hline
			\textbf{Centromere}        & 208                                                & 149                                              & 149                                               & 149                                             & 2741                                                    & -                                                     \\ \hline
			\textbf{Cytoplasmic}       & 58                                                 & 51                                               & -                                                 & -                                               & -                                                       & -                                                     \\ \hline
			\textbf{Nuclear Membrane}  & -                                                  & -                                                & -                                                 & -                                               & 2208                                                    & -                                                     \\ \hline
			\textbf{Golgi}             & -                                                  & -                                                & -                                                 & -                                               & 724                                                     & -                                                     \\ \hline
			\textbf{Total}             & \textbf{721}                                       & \textbf{734}                                     & \textbf{905}                                      & \textbf{979}                                    & \textbf{13596}                                          & \textbf{-}                                            \\ \hline
	\end{tabular} }
\end{table*}

Neslihan et. al.~\cite{bayramoglu20} analyzed the effects of data pre-processing, augmentation, and pre-training within a CNN based framework. They demonstrate that an additional real data-augmentation is incredibly helpful to achieve high performance. In~\cite{yang2013}, authors trained a filter bank by using Independent Component Analysis (ICA) for feature generation. Wiliem et. al~\cite{wiliem20} proposed a system that comprise of a dual-region codebook-based descriptor combined with the Nearest Convex Hull Classifier. Richa et. al~\cite{nigam20} proposed a framework which combines Laws features with two level SVM classifier coupled with posterior class probabilities. In ~\cite{strandmark}, extracted large bunch of features using thresholding at different intensity levels and used random forest for classification. Nosaka et al.~\cite{Nosaka}, utilized a novel descriptor, namely rotation invariant co-occurrence  among adjacent LBP (RIC-LBP) in order to extract texture information. RIC-LBP achieves robustness against local rotations on cell images, however it is not able to capture global rotation. To deal with global rotation they synthesized additional training images by rotating the original training images and made use of multi-class SVM classifier for both the original and synthesized training images. A method that combined following image descriptors: Haralick features, Local Binary patterns, SIFT, surface description and a granulometry-based descriptor was proposed by Stoklasa et al.~\cite{Stoklasa}. The K-NN classifier was used for classification and four k-nearest neighbor were computed using custom aggregated distance function that combines different descriptors. Liu et al.~\cite{Lingqiao} introduced a multi-projection-multi-codebook strategy which builds multiple descriptors for a local patch and creates multiple codebooks to generate multiple pooled vector for an image. The main idea of this paper was to learn the appropriate descriptor from the image data itself. Partial least square (PLS) was used to select most representative descriptor for an image. Shen et al.~\cite{Shen} combined the intensity order pooling based feature with bag of words (BOW) framework. Multisupport Region Order-based Gradient Histogram (MROGH) employed a local coordinate system to calculate gradient feature and intensity order for feature pooling. Spatial pyramid matching was introduced to BOW in order to incorporate spatial information. As no orientation estimation was required, MROGH is more robust than SIFT against rational variance. In a quite different manner, Di Cataldo et al.~\cite{Cataldo} proposed an automated solution for HEp-2 staining pattern classification  based on Subclass Discriminant analysis approach. Firstly they obtained an optimal set of features that combines morphological descriptors with global and local textural analyses. Second, feature selection coupled with SDA was shown to  effectively cope with the high within-class variance and at the same time allows compact set of image attributes. Wiliem et al.~\cite{Wiliem} introduced novel framework composed of Cell Pyramid Matching (CPM) descriptor with Multiple kernel learning. The idea of CPM was drawn from spatial pyramid matching (SPM) and Dual Region (DR).The above discussed methods were bench-marked with the datasets proposed by ICPR-2012 and SNHEp-2.
In a similar perspective as ours, of utilizing morphologically motivated features, Gennady et al.~\cite{Gennady} has made use of simple features from the cell regions such as number, size, location, shape  and used SVM classifier for the classification of images using obtained set of image features.  However, this is a relatively limited study, also carried out on a smaller dataset (ICPR-2012~\cite{Saggese}).

We now discuss more recent works which use the larger dataset from ICPR-2014. In ~\cite{cascio2016}, authors adopted non-standard pipeline for classification, where they used combined set of complementary process instead of single. For this purpose, they analyzed various preprocessing algorithms and selected one which provided best performance in terms of class accuracy for each block. Shahab et. al~\cite{ensafi2016} proposed a super-pixel based system to classify HEp-2 cell images. In ~\cite{nanni2016}, various texture descriptors were fused, and an ensemble of SVMs combined with sum rule was used for classification. Xian et. al~\cite{han2016} proposed a local ternary pattern, by introducing the Rotation Invariant Co-occurrence Weber-based Local Ternary Pattern (RICWLTP). The RICWLTP descriptor incorporates the contexts of spatial and orientation co-occurrences among adjacent Weber-based local ternary patterns. Diego et al.~\cite{gragnaniello2016} used Scale-Invariant Descriptor (SID) to classify HEp-2 cell images. The bag-of-word model with soft assignment through Gaussian weights, with linear SVM employed for classification. In~\cite{Zhimin,rodrigues2017,jia2016}, authors utilized CNN based framework along with various preprocessing for HEp-2 cell classification. Nanni et. al.~\cite{L} used an optimal ensemble of texture and morphological descriptor and performed classification using SVM with Radial basis function (RBF), whereas Ilias et. al.~\cite{T} adopted a combination of morphological features with bundle of textural descriptors and used SVM with linear kernel for classification. Instead of morphological features,~\cite{Theodorakopoulos} proposed an algorithm to extract statistical features. For this purpose, they used two statistical descriptors, one is Fuzzy Size Zone Matrix (FSZM, a fuzzy version of GLSZM) and another is Multi-resolution Local Binary Patterns (MLBP). The FSZM estimates the bivariate conditional probability density function of the image distribution values, whereas the latter is a gray level invariant technique in which histograms of pixels characterize the texture.  In~\cite{Ensafi}, Bag of Words model based on sparse coding was proposed. They used scale-invariant feature transform (SIFT) and speeded-up robust features (SURF) features with sparse coding and max pooling. Diego et. al.~\cite{Gragnaniello} utilized a biologically-inspired dense local descriptor for characterization of cell images.  In~\cite{Donato}, structured classification process based on k-Nearest Neighbors classifiers (KNN) was introduced. They extracted a large pool of features which characterize the staining pattern, in combination with different set of complementary process to discriminate each class from all other classes. Chandran et al.~\cite{Chandran} exploited first, second and third order statistics of pixel intensity distributions over predefined regions to extract features which contain texture information of different regions.

\begin{table*}[h!]
	\caption{Summary of related work.}
	\label{table2} 
	\centering
	\resizebox{13.80cm}{!}{
		
		\begin{tabular}{lllll}
			\hline
			\textbf{Approach} & \textbf{Preprocessing} & \textbf{Descriptor} & \textbf{Classifier}  \\ \hline
			\multirow{3}{*}{\textbf{Nanni et al.~\cite{L}}} & \multirow{3}{*}{}      & \multirow{3}{*}{\begin{tabular}[c]{@{}l@{}@{}}Ensemble of 5 feature sets:  Multiscale pyramid(PLBP)\\RIC-LBP, Local configuration pattern(LCP), Extended LBP(ELBP)\\ and Strandmark morphological features (STR)\\\end{tabular}}     & \multirow{3}{*}{Radial basis function SVMs}\\
			&                        &                     &                     &                   \\
			&                        &                     &                     &                   \\
			\multirow{3}{*}{\textbf{Ilias et al.~\cite{T}}} & \multirow{3}{*}{Intensity normalization}      & \multirow{3}{*}{Combination of morphological features and bundle of textural descriptors}   & \multirow{3}{*}{SVM with linear kernel}   \\
			&                        &                     &                     &                   \\
			&                        &                     &                     &                   \\
			\multirow{3}{*}{\textbf{Shahab et al.~\cite{Ensafi}}} & \multirow{3}{*}{}      & \multirow{3}{*}{SIFT and SURE features}   & \multirow{3}{*}{SVM with linear kernel}   &  \\
			&                        &                     &                     &                   \\
			&                        &                     &                     &                   \\
			\multirow{3}{*}{\textbf{Diego et al.~\cite{Gragnaniello}}} & \multirow{3}{*}{}      & \multirow{3}{*}{Dense scale and rotation invariant descriptor}   & \multirow{3}{*}{SVM with linear kernel}   \\
			&                        &                     &                     &                   \\
			&                        &                     &                     &                   \\
			
			\multirow{3}{*}{\textbf{Cascio et al.~\cite{Donato}}} & \multirow{3}{*}{Intensity normalization}     & \multirow{3}{*}{\begin{tabular}[c]{@{}l@{}}108 features based on : features intensity,\\Geometry, Shape-Morphology and Descriptors are extracted \\\end{tabular} }   & \multirow{3}{*}{K-NN}  \\ 
			&                        &                     &                     &                   \\
			&                        &                     &                     &                   \\
			
			\multirow{3}{*}{} & \multirow{3}{*}{}      & \multirow{3}{*}{}   & \multirow{3}{*}{}    \\
			&                        &                     &                     &                   \\
			\multirow{3}{*}{\textbf{Siyamalan et al.~\cite{Manivannan}}} & \multirow{3}{*}{Intensity normalization}      & \multirow{3}{*}{\begin{tabular}[c]{@{}l@{}}Sparse encoding of texture features with cell pyramids,\\Capturing spatial and multi-scale structure \\\end{tabular} }   & \multirow{3}{*}{Ensembles of SVMs}   \\
			&                        &                     &                     &                   \\
			&                        &                     &                     &                   \\
			\multirow{3}{*}{\textbf{Siyamalan et al.~\cite{S}}} & \multirow{3}{*}{Intensity normalization}      & \multirow{3}{*}{\begin{tabular}[c]{@{}l@{}}Four types of local descriptor : Multiresolution local pattern, Root-SIFT,\\Random projections, Intensity histograms \\\end{tabular} }   & \multirow{3}{*}{Ensemble of SVMs }  \\
			&                        &                     &                     &                   \\
			&                        &                     &                     &                   \\ \hline
			
	\end{tabular}}
	
\end{table*}

Apart from different features and standard classification strategies, ensemble learning has also attracted much attention for many applications in medicine and biology. Ensemble learning is concerned with mechanisms to combine the results of a number of classifiers. With the similar work on  boosting ~\cite{Robert} and Adaboost algorithm ~{\cite{Schapire}} which show that a strong classifier can be generated by the combination of several weak classifiers, the idea of ensemble learning was first introduced in the late of 1970's. Siyamalan et. al.~{\cite{Manivannan}} presented an  ensembles of SVMs based on sparse encoding of texture features with cell pyramids, capturing spatial, multi-scale structure for HEp-2 cell classification and reported mean accuracy. A method, based on multiple types of local feature which uses two -level pyramid to retain some spatial information was proposed by~{\cite{S}}. In this, they utilized ensemble of linear support vector machines for classification of cell images. In different domain such as gene classification, breast cancer diagnosis etc.  ensemble learning has also been utilized. In~{\cite{Zhang}}, a cascade system which utilizes classifier ensembles of Support Vector Machine and  Multi-layer perception was proposed for breast cancer diagnosis. For the same application, an ensemble of One-class SVM was also introduced by~{\cite{Zhang2}}. In ~{\cite{Zhimin}}, authors utilized deep learning based framework for cell classification. 

The above discussed methods were bench-marked with the dataset proposed by  ICPR-2014~\cite{Foggia}. Table~\ref{table2} provides a summary of some frameworks proposed for HEp-2 cell classification over the last five years. While the approaches discussed above have clearly helped the progress in the area of HEp-2 cell classification, most of these methods mainly focus on using sophisticated feature extraction methods (SIFT, SURF, LBP, HOG etc.), and the studies are with single standard off-the-shelf classification framework. 

\textbf{Summary of contribution:}
The main contributions of this paper are: (1) employing simplistic and visually more interpretable feature definitions separately for each class (class-specific), (2) Considering that the problem consists of a few classes, various classification frameworks are evaluated with different feature sets.
(3) demonstrate the superiority of class-specific features across all classification strategies. (4) experimental analysis including, (a) classification of low contrast images which are important in detecting the onset of the diseases, (b) experimentation with a small amount (only 40\%) of data for training. To the best of our knowledge, such a study and analysis is not yet reported for HEp-2 cell classification and we believe that this is a new direction which would benefit computer aided diagnosis.

\textbf{Extension over our earlier work:} The present work aims to significantly expand upon our earlier work~\cite {gupta2016}. In ~\cite {gupta2016}, we considered only of the classification strategy with relatively limited number of combinations of class-specific features. This work reports the results utilizing various combination of features, and various feature sets that involves larger dimensionality of features. Furthermore, to demonstrate the superiority of proposed features, various classification frameworks which involves traditional structure, two-stage hierarchies, and random forests are also evaluated. The present work also involves a larger set of experiments and results including comparisons with different classification strategies and with some contemporary approaches. Finally, we also provide a much more elaborate coverage for most sections including related work, feature description, classification, experimental results etc.

The rest of the paper is organized as follows. In the next section, we provide the details of dataset that we employ. Section 3, illustrates various classification frameworks along with class-specific and traditional features, and a brief description of support vector machine (SVM) and random forests. In section 4 and 5, the experimentation parameter and  results are discussed. Section 6 concludes the paper.

\section{Dataset}
The proposed approach has been evaluated using publicly available dataset from the \textbf{ICPR} 2014 HEp-2 cell image classification contest, comprising more than 13,000 cell images~\cite{Hobson}. The dataset has been used in many related works on HEp-2 cell classification. It consists of six classes termed as: Homogeneous (H), Speckled (S), Nucleolar (N), Centromere (C),  Nuclear Membrane (NM), Golgi (G). In the dataset, each class consist of positive and intermediate images. The intermediate images are generally lower in contrast compared to the positive images. The dataset also includes mask images which specify the region of interest of each cell image. The details of dataset are given in Table~\ref{table1}.

\begin{table}[h!]
	\centering
	\caption {ICPR-2014 HEp-2 cell database.} 
	\label{table1}
	\medskip
	\setlength\extrarowheight{3.3pt} 
	\resizebox{10.03cm}{!}{
		\begin{tabular}{|c|c|c|c|}
			\hline
			\textbf{Classes}               & \textbf{Positive Images} & \textbf{Intermediate Images} & \textbf{Total} \\ \hline
			\textbf{Homogeneous (H)}       & 1087                     & 1407                         & 2494           \\ \hline
			\textbf{Speckled (S)}          & 1457                     & 1374                         & 2831           \\ \hline
			\textbf{Nucleolar (N)}         & 934                      & 1664                         & 2598           \\ \hline
			\textbf{Centromere (C)}        & 1378                     & 1363                         & 2741           \\ \hline
			\textbf{Nuclear Membrane (NM)} & 943                      & 1265                         & 2208           \\ \hline
			\textbf{Golgi (G)}             & 343                      & 375                          & 724            \\ \hline
	\end{tabular}}
\end{table}

\section{Proposed framework}
This section describes the proposed framework, that can be decomposed into following parts: (1) Preprocessing, (2) Feature extraction, (3) Feature normalization, (4) Classification strategies.

\subsection{\textbf{Preprocessing}}
As mentioned earlier, the dataset (Table~\ref{table1}) contains two kinds of images for each class viz. positive and intermediate. Prepossessing is required for enhancing intermediate-quality images, which have low contrast in comparison to the positive images. However, as one would typically, not know if an image is positive or intermediate during the test phase, we apply preprocessing for all images. Hence, in order to increase the contrast of cell images, a rescaling of the images between [0,1], followed by a standard $\gamma$ transformation is performed on each cell image before extracting the features. 
We note that the value of $ \gamma $ is not fixed for all features. It varies, depending on the feature definition, because a single value of $ \gamma $ is not suitable for all feature definitions.

\subsection{\textbf{Feature extraction}} 
This subsection talks about the class-specific (CS) and standard texture (ST) feature sets. As CS are the proposed features, they are described in detail while for texture features appropriate reference is provided.

\subsubsection{\textbf{Class-specific features}} 
As mentioned in the introduction, in the existing approaches, commonly available features for describing morphology and texture are extracted. These approaches typically follow the conventional strategy of extracting the same features across all classes  without explicitly taking into account the visually observed traits of each class. 

The basic intuition behind our approach is that instead of using a single feature to discriminate each class from all other classes, we utilize visual characteristics of classes to formulate class-specific features. This can eventually contribute to a class-specific cell classification, where the features specific to different classes are used. Further, class-specific features are rather limited in number and help to reduce the feature space which allows efficiency in learning and representation. The characteristic definitions of fluorescence patterns considered in this work, based on its visual traits, can be stated as~\cite{Foggia} :
\begin{enumerate}
	
	\item\textbf{Homogeneous}: a uniformly diffused fluorescence covering the entire nucleoplasm sometimes accentuated in the nuclear periphery.  
	
	\item\textbf{Speckled}: these pattern have two sub categories (which, however, are not separately labeled in the dataset)
	
	- Coarse speckled: density distributed, various sized speckles, generally associated with larger speckles, throughout nucleoplasm of inter-phase cells; nucleoli are negative.
	
	- Fine speckled:fine speckled staining in a uniform distributed, sometimes very dense so that an almost homogeneous pattern is attained; nucleoli may be positive and negative.
	
	\item\textbf{Nucleolar}: characterized by clustered large granules in the nucleoli of inter phase cells which tend towards homogeneity, with less than six granules per cell.   
	
	\item\textbf{Centromere}: characterized by several discrete speckles (40-60) distributed throughout the inter phase nuclei. 
	
	\item\textbf{Nuclear Membrane}: A smooth homogeneous ring like fluorescence of the nuclear membrane.
	
	\item\textbf{Golgi}: Staining of polar organelle adjacent to and partly surrounding the nucleus, composed of irregular large granules. Nuclei and nucleoli are negative.
\end{enumerate}

The above specification of classes suggests that there is a unique visual feature for each class for its characterization. For instance, in most existing approaches, the features, no matter how sophisticated,  were extracted using existing mask information from inside the nucleus where there is least information for the Golgi class. Similarly, for the Nuclear Membrane class the important representative information is found in ring around the nucleus boundary. In the proposed approach, features are extracted from specific region called region of interest (ROI) which contains the useful information for a particular class, computed using the mask images. Except variance and mean, all proposed features for each class are extracted from binary image $I_{b}$, which is obtained after thresholding the processed image. The value of threshold $ (T) $ is chosen experimentally. After thresholding, any connected group of pixels (foreground pixels) is referred as a object. In addition to the location of ROI, extracted feature definitions also depend on visual implying characteristics. Features which are specific to particular class are described as follows:

\begin{enumerate}
	\item\textbf{Homogeneous class (H)}: The given explanation for the Homogeneous class reflects that, an entirely stained nucleus whose intensity varies depending on number of antibodies in the serum, is a  prominent observation which can be used for its characterization. The homogeneously stained nucleus is transformed into single object after thresholding, which entails largest object's size and maximum perimeter. As this class shows the presence of single object in a single cell, due to this entire mask is considered as ROI. Thus, the image features  which capture the uniqueness of this class are given as follows: 
	
	\begin{enumerate}
		\item Maximum object Area (MOA): Area of object having a maximum size.
		\item Area of connected components (ACC): This feature measures the number of white pixel inside the nucleus. As this class has an entirely stained nucleus, the value of ACC will be high.
		\item Maximum Perimeter (MP): As the name suggest, it account for the maximum periphery of object inside the nucleus.
	\end{enumerate}
	All proposed features are computed from thresholded image (Fig.~\ref{fig:two}(e)). Due to the presence of a single object in homogeneous class, the value of defined features will, ideally, be high. 
	
	\begin{figure}[h!]
		\centering
		\resizebox{10.03cm}{!}{
			\includegraphics[width=\columnwidth]{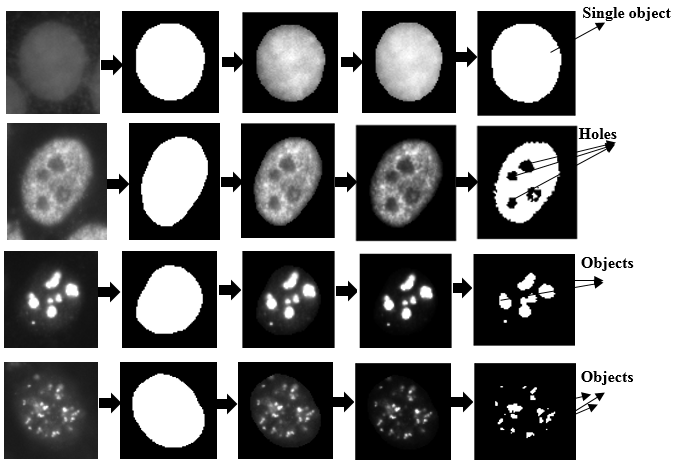}}
		\caption{\label{fig:two}Column wise : (a) Original image of homogeneous, speckled, nucleolar, centromere class. (b) Existing mask. (c) Image after multiplying (a) and (b). (d) Pre-processed image. (e) Thresholded Image.}
	\end{figure}
	
	\item \textbf{Speckled class (S)}: An eminent feature that is unique for this class is negatively stained nucleoli located inside a positively stained nucleus. The nucleoli appear in the speckled classes of image as round dark holes inside a bright nucleus circle (Fig.~\ref{fig:two}(a)). Thus, image features which help to recognize the  nature of various size holes are defined as follows : 
	\begin{enumerate}
		\item Holes Inside the object (HN): Total number of holes inside the cell.
		\item Hole Area (HA): Total number of black pixel in a cell.
		\item Euler Number (EN): This number gives the relation between number of object and number of holes.
		$E_{n} = Nc - Nh.$
	\end{enumerate}
	As the holes are scattered inside the nucleus, the original mask is considered as the ROI mask.

	
	\item\textbf{Nucleolar (N)}: An intense staining of nucleoli could be observed in nucleolar pattern.  Although there is a single nucleolus present in cell nucleus but at the time of cell-division, more than one nucleolus may be present due to binary fission. It is noticed that the number of nucleolar patterns varies from 2 to 6. This constitute an important observation and widely assists in accurate characterization. The third row of Fig.~\ref{fig:two} illustrates the whole process of feature extraction for nucleolar class. The following features are proposed based on above description to exhibit uniqueness of this class.
	\begin{enumerate}
		\item Maximum object Area (MOA): Area of object having a maximum size.
		\item Average object Area (AOA): Average area of objects.
		\item Connected Component (CC): Total number of objects in the cell.
	\end{enumerate}
	
	
	\item\textbf{Centromere class (C)}: The centromere pattern is distinguished by uniformly stained nuclei with fine speckles distributed in the nucleoplasm of interphase cells~\cite{Cabiedes}. The size of stained cell nucleoli for this class is somewhat smaller than the nucleolar class, which affirms that smallest size of object will be obtained here. The number of speckles present in cell nucleus are associated with the number of chromosomes present in nuclear region (40-60).  Thus, the image features  which capture the uniqueness of this class are same as Nucleolar class (MOA, AOA, CC).
	
	Chromosomes and nucleoli are uniformly distributed throughout the entire nucleus in nucleolar, centromere class respectively. Owing to this entire mask is considered as ROI to extract the useful information. 
	
	
	\begin{figure}[h!]
		\centering
		\resizebox{7.2cm}{!}{
			\includegraphics[width=\columnwidth]{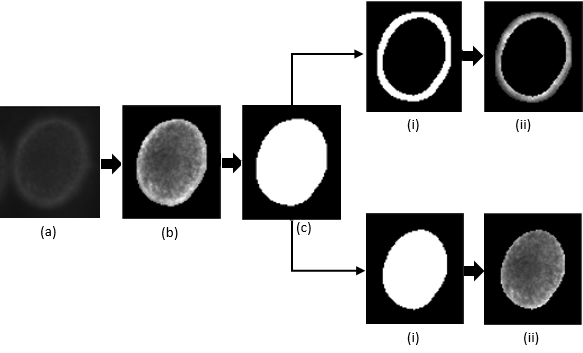}}
		\caption{\label{fig:three}(a) Original image of NM class. (b) Processed image. (c) Existing mask. (d) Upper row: (i) Proposed mask for BAR. (ii) ROI for BAR. (e) Lower row: (i) Proposed mask for IAR. (ii) ROI for IAR.}
	\end{figure}
	
	\item\textbf{Nuclear Membrane class (NM)}:  A ring like structure on the nucleus boundary  and  low intensity of image information in the intra nuclear space are  unique features which aid in differentiating NM class from other classes. This indicates  that useful information can be found in a ring which is centered on the boundary and inside area. To confine the  specific region, for extracting ring like structure and low intensity, the following region of interest (ROI) masks are defined: (1) To recognize the ring structure, the ROI mask is a ring with a width of $k_{b} $ pixels centered on the boundary of original mask. (2) To capture low intensity inside nucleus, inside region of original mask by leaving out  $k_{i} $ pixel from cell boundary is considered as ROI mask. Fig.~\ref{fig:three} depicts the overall process to make ROI masks which are used for extracting unique information. To demonstrate the unique structure of NM class using ROI masks, following features are defined~\cite{Gupta}:
	\begin{equation}
	Area~Ratio~(AR) = \displaystyle{\frac{\text{No.}~ \text{of}~1'\text{s} ~ \text{in} ~ \text{the}~\text{ROI} ~ \text{in}~ I_{b}}{\text{Total}~\text{no.}~\text{of}~\text{pixels}~\text{in}~\text{the}~\text{ROI}}}
	\end{equation}
	
	
	\begin{enumerate}
		\item Boundary Area Ratio (BAR): It is a area ratio in boundary ring mask.
		\item Inner Area Ratio (IAR): It is a area ratio in inner mask.
		\item Eroded Area Connected Component (EACC): It gives the number of white pixels in inner mask.
	\end{enumerate}
	
	\item\textbf{Golgi class (G)}: The above description of Golgi class suggests that useful information can be extracted in the region inside and outside of the nucleus. Thus, based on this simple observation the following ROI masks are defined to consider outward region of nucleus and inside low intensity : 
	the inside region of original mask leaving out $k_{i} $ pixel from cell boundary (inner mask)  is used to capture low intensity inside cell nucleus~\cite{Gupta}. Fig.~\ref{fig:four} illustrates the overall process used to form ROI masks. The following features are proposed to extract defined characteristics~\cite{Gupta}:
	
	\begin{figure}[h!]
		\centering
		\resizebox{7.2cm}{!}{
			\includegraphics[width=\columnwidth]{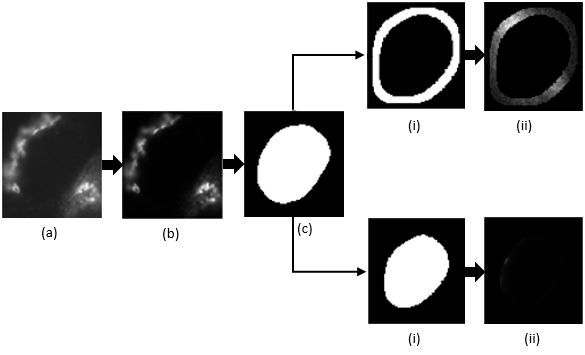}}
		\caption{\label{fig:four}(a) Original image of golgi class. (b) Processed image. (c) Existing mask. (d) Upper row: (i) Proposed mask for OAR. (ii) ROI for OAR. (e) Lower row: (i) Proposed mask for IAR. (ii) ROI for IAR.}
	\end{figure}
	
	\begin{enumerate}
		\item Outer Area Ratio (OAR): It is a area ratio computed using outer mask.
		\item Eroded Area Connected Component (EACC): It gives the number of white pixels in inner mask.
		\item  Average object distance (AOD): In this class the antibodies targets are located outside of the nucleus while in other classes it is closer to the center of image. The feature Average Object Distance (AOD) is defined  to describe whether targets are outside of the nucleus or closer to center. AOD is average of the shortest distances from each object's pixel to the image boundaries~\cite{Gennady}.
		
	\end{enumerate}
\end{enumerate}

\subsubsection{\textbf{Standard traditional features~\cite{strandmark}}}  
We now briefly describe some standard texture features which we employ to compare with the proposed CS features.  Features which include morphology (like Number of objects, Area, Area of the convex hull, Eccentricity, Euler number, Perimeter), texture (like Intensity,  Standard deviation, Entropy, Range, GLCM) are extracted at 20 intensities equally spaced from its minimum to its maximum intensity. We term these standard features as traditional features. These extracted features are same as the features which were used in~\cite{strandmark}. However, in~\cite{strandmark}, these features were applied on much smaller dataset.

\subsubsection{\textbf{Combined feature set}} 

Class-specific features (EACC, BAR, OAR, IAR, AOD) which were not present in standard traditional set are added. This feature set helps to see the role of proposed class-specific features in HEp-2 cell image classification.

\subsection{\textbf{Feature normalization}}

Feature normalization neutralize the effect of different scales across features. The main advantage of normalization (scaling) is to avoid attributes in greater numeric ranges dominating those in smaller numeric ranges.  It has been seen that, feature normalization would increase the accuracy, especially given the large differences in features. There are various ways to normalize the feature vector.  In the present work, we have used mean and standard deviation (z-score) to normalize the features.  The definition of z-score is given as follows:

\textbf{Z score:} Mean and standard deviation are computed for each column separately. The values in the column are transformed using the following formula:

\begin{equation}
Z =\dfrac{X-mean(x)}{stdev(x)}
\end{equation}

The validation and testing data, are normalized using  mean and standard deviation calculated from training data. In equitation \textit{X} is a feature vector (testing/validation) which needs to normalize, and \textit{x} is a training feature vector.

\subsection{\textbf{Classification Strategies}}
In this subsection, we discuss the classification strategies which we have used for the proposed study, and their salient aspects. In most of the strategies considered in this work, we use the support vector machine (SVM), and decision tree as a base classifier. 

\subsubsection{\textbf{One-vs-one}}
The one-vs-one strategy is one of the default ways to solve the multi-class classification problem by support vector machine (SVM). SVM, fundamentally a binary classifier, handles the multi-class classification by decomposing the problem into multiple binary classification problems and using a decision strategy. Here, the same features along with same parameter setting are utilized by each binary block.

\begin{figure*}[t!]
	\centering
	\begin{tabular}{c c}
		\hspace{-0.3cm}
		\includegraphics[height=120pt, width=160pt]{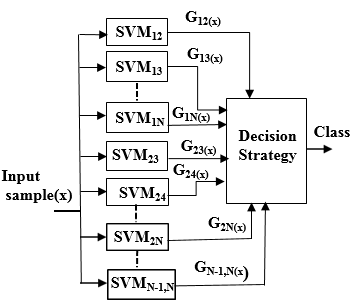}&
		\hspace{-0.15cm}
		\includegraphics[height=120pt, width=160pt]{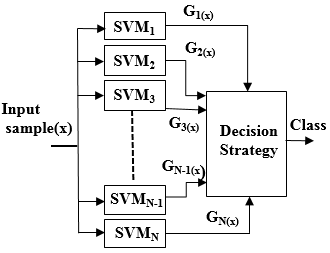}\\(a)&(b)
	\end{tabular}
	\caption{\label{fig:five}(a) One-vs-one approach  (b) One-vs-rest approach.}
	\label{}
	\vspace{-0.1in}
\end{figure*}

This strategy constructs SVM  for every pair of classes to form a boundary between their regions. The max-win strategy is used to determine the class of a test pattern. Typically, to solve N-class problem, SVM builds N(N-1)/2 one versus one models using training data~\cite{Chang}. Fig.~\ref{fig:five}(a) shows the architecture of one vs. one strategy. Following this, each binary classification casts votes for all the test samples, finally the test sample is designated to be in a class with the maximum number of votes. In the case of a tie (among two classes with an equal number of votes), it selects the class with the highest aggregate classification confidence by summing over the pair-wise classification confidence levels computed by the underlying binary classifiers. From the perspective of ensemble classification, this approach can be considered as an ensemble of multiple binary classifiers. In this strategy, the same high dimensional feature (consisting of a stacking of all the scalar features) and parameters are used across all one-vs-one classification blocks.

\subsubsection{\textbf{Two-stage strategies}} 

We note that the above discussed framework is essentially a scheme to decompose the multi-class classification into the multiple binary classification problems. Clearly, we can also have variations to such a strategy which uses the binary classifiers differently. 
In this subsection, we discuss the  strategies which are designed in two stages while using multiple binary classifiers. These stages are sequentially connected i.e. output of first stage is feeded to the second stage. The first stage mainly focuses on achieving high true positive rate for all classes. This is due to the observation that in health-care diagnostic screening systems, false negatives are considered to be more detrimental than false positives, as the latter can be mitigated in later stages of diagnosis after the basic screening. However, not detecting a particular true abnormality (false negative) can be dangerous. The focus of the second stage is to reduce the false positives.

We firstly discuss the ways by which first stage can be structured. After designing the first stage,  under the heading \textbf{Second stage for resolution of ambiguities}  we discuss the methods to resolve the ambiguities (from first stage). Any of these methods  can be used in conjunction with first stage.

\begin{itemize}
	\item \textbf{One-vs-rest approach:} This essentially consists of a set of various binary verification blocks, one for each class where each block decides whether a test pattern belongs to a particular class or not. The overall classification involves N such verification blocks. This is feasible for problems which involve a small number of problems. 
	
	This strategy constructs one SVM classifier per class to form a boundary between the region of the class and the regions of the other classes. For each classifier, the class is fitted against all the other classes. Fig.~\ref{fig:five}(b) details the  one vs. rest strategy. Similar to the one-vs-one classifier, in this case, all such one-vs-rest verification blocks employ the same features and parameters. A test pattern is classified by using winner-takes-all strategy (where, ideally, a test sample should be assigned correctly to only one block). In the case of a tie (where test sample may be designated as positive in more than one classification blocks), we follow some strategies in a second-stage, as we discuss below. 
	
	We note that, from a system perspective, 
	this sort of analysis can be considered similar to differential diagnosis followed by clinicians, wherein the ambiguity among diagnosis hypothesis is not completely resolved at an initial stage, but is significantly reduced. Thus, a CAD system can, in principle, independently use the output of the first stage, for `shortlisting' the ambiguous classes for further diagnostic procedures.
	
	\hspace{0.5cm}
	
	\item \textbf{Hierarchical strategy:} 
	As a variant of the verification strategy discussed above, the utilization of class-specific features and leveraging the fact that problem involves few classes (6 in this case), each verification block for a class may be structured in a small number of hierarchical stages, where at each stage, 
	a binary one-vs-one classification is carried between two classes (Fig. 7(a)). Importantly, in each of the one-vs-one sub block, one can have flexibility of choosing  a small subset of 3 to 6 scalar class-specific features, based on the two classes involved in that one-vs-one block. This subset of features is chosen via an exhaustive validation experiments during the training - validation phase.
	\hspace{0.5cm}
	
	
	\item \textbf{Hierarchical strategy with common features:}
	
	This strategy is designed in same way as above (hierarchical structure). The only difference between these two, is that the feature set used at different stages of each verification block involves all features of one type (e.g. class-specific or texture or combined). However, note that the parameter setting is allowed to change for different hierarchical stages in a verification block.
	
	\item \textbf{Second stage for resolution of ambiguities:} As mentioned earlier, using just the above strategies can result in some samples 
	being recognized by more than one verification block, in the first stage. 
	Overall, this can result in some false-positives, which can be reduced by resolving such ambiguities. Considering the ambiguities, we use the following naming conventions. The true positive samples at the end of the first stage (Overall True Positive) (OTP)) are divided as certain true positives (CTPs), which are accepted by only the correct block, and ambiguous true positives (ATPs), which are accepted as true also by some of other classification blocks. 
	The latter will contribute to the false positives (OFP). We perform resolution of ambiguities as discussed next, for the above cases.
	
	\begin{itemize}
		\item \textbf{One-vs-rest (using SVM score):} While the SVM classifier also yields a real-valued classification score, typically the sign of such a score is used for the decision. To mitigate the confusion among ambiguous samples (ATPs) from the first stage, we use the real-valued SVM score. SVM score is a distance of input sample from decision boundary given by support vector machine (SVM).  
		For the ambiguous true positives (ATPs), we calculate the distance (SVM-score) for each verification block. For instance, if a given test sample is classified as positive in 2, 3  and 4 classes, we calculate the distance given by blocks which were designed for these classes, and compare distances. The class for which the distance is high (implying that the sample is confidently classified), the sample is assigned only to that class.
		
			\item \textbf{Hierarchy with common features (using SVM score):} As in the one-vs-rest case, here too, we used SVM-score to resolve ambiguities among samples obtained from the first stage. However, unlike one-vs-rest approach where only one score is generated corresponding to each block, here five such scores are produced for each block, as each block is made up of five one-vs-one blocks. The final score for each block is average of five sub-blocks. The rest of the procedure is same as what we have applied in one-vs-rest case. As we already use high dimensional features in first stage, we cannot use the strategy involving the high-dimensional features as done in the hierarchical approach.
			
		\item \textbf{Hierarchical approach (Second-stage with independent binary blocks):}
Here, we observe that there is scope to incorporate more refined features at second stage as first stage utilized only low-dimension features.  Fig.~7(b) shows the classification model utilized for the second stage. 
		
	More specifically, at the second stage, a high dimensional feature set (consisting of either all class-specific features, standard scalar texture features, or their combination) is used in order to mitigate the confusion among samples. This model comprises 15 binary blocks where each block utilizes the same feature set with different parameter setting. Unlike the first, stage, here we employ only 15 binary classification blocks as all of them involve the same features. However, importantly, not all 15 classifiers are used for all samples. The ATP samples will pass through the respective classification with respect to which there is an ambiguity.  For example, for H v/s S block the input will be the samples which were accepted by both verification blocks of H and S in the first stage. We note that for most of the samples, only one block is selected, as most ATP samples have an ambiguity among only 2 classes. Less than 3\% of ATP samples have ambiguity among more classes. 
		
		A sample with ambiguity among more than 2 classes for example, if sample is recognized by three verification block H, C and NM, it is passed through all binary blocks involving these classes.
The sample will be assigned to the class for which all corresponding block classify it correctly. 

	\end{itemize}
	
\end{itemize}

\begin{figure}
	\centering
	\begin{tabular}{@{}c@{}}
		\includegraphics[width=1\linewidth]{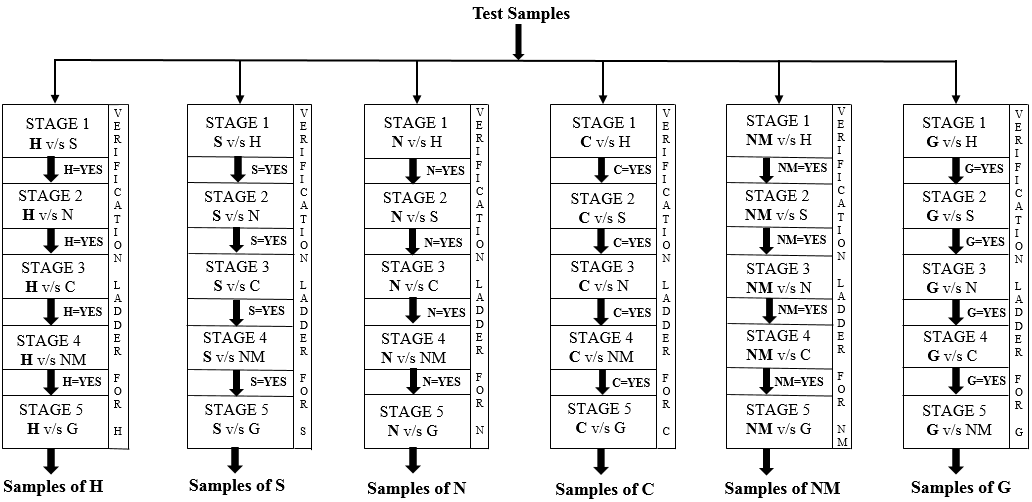} \\[\abovecaptionskip]
		\small (a)
	\end{tabular}
	
	\begin{tabular}{@{}c@{}}
		\includegraphics[width=1\linewidth]{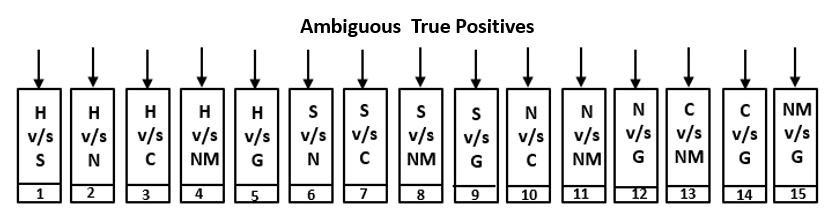}\\[\abovecaptionskip]
		\small (b)
	\end{tabular}
	\caption{(a) Two-stage classifier: First stage. (b)Two-stage classifier: Second stage.}\label{fig:six}
\end{figure}

\subsubsection{\textbf{Ensemble frameworks}}
All the above discussed methods can be seen as an ensemble of SVM classifier. Hence, in the same spirit, in this study we also include frameworks which are ensemble of decision trees, and which are also considered as popular contemporary classification paradigms.

\begin{figure}[h!]
	\centering
	\resizebox{10.03cm}{!}{
		\includegraphics[width=\columnwidth]{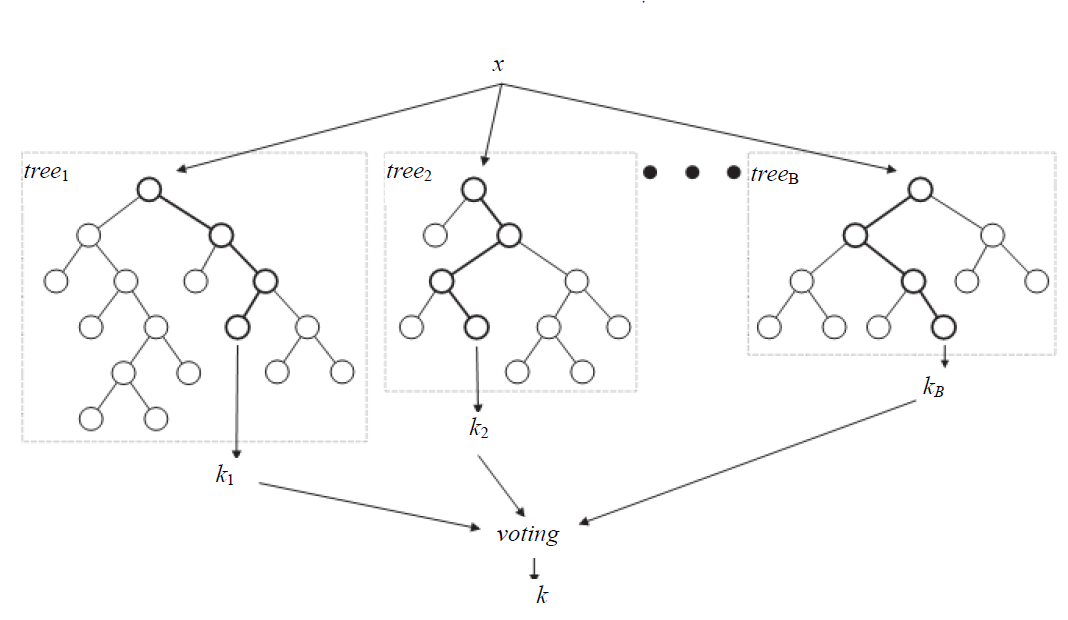}}
	\caption{\label{fig:nine}Random Forest~\cite{breiman}.}
\end{figure}

\begin{itemize}
	\item \textbf{Bagging:} Bagging (short for Bootstrap Aggregation) algorithm is an ensemble-based algorithm. In bagging, diversity among base classifiers  is attained by utilizing bootstrapped replicas of the given training data. For example, given a training dataset \textit{S} which consists \textit{N} samples, bagging simply trains \textit{T} independent classifiers where each one is trained on sample set which is selected using \textbf{sampling with replacement procedure}.
	\begin{itemize}
		\item \textbf{Random forest (RF):} Random Forest~\cite{breiman}, is one of the modification of bagging where decision tree is used as base classifier. Along with bootstrapped replicas, random sub-sampling is also used to include diversity at feature level i.e  each node of a tree is presented with a new random selection of features. After training, predictions of all base classifiers are combined by taking a simple majority vote. Fig.~\ref{fig:nine} shows the general structure of random forest.

		\item \textbf{Random Uniform Forest (RUF): }RUF~\cite{ciss} is a another modified version of bagging where random uniform decision trees, which are unpruned and binary are used as base classifier. To create partition from each node, random `cut-points' is generated assuming an uniform distribution for each candidate variable. An important motivation of this algorithm is to  
		get tress which are less correlated, to allow a better analysis of variable importance.

	\end{itemize}
	\item \textbf{Boosting:} Boosting is an iterative approach where the training dataset for each subsequent classifier increasingly focuses on  mis-classified instances generated by previous  classifiers, different from bagging where each instance has equal chance of being in each training dataset. To combine the predictions given by subsequent classifiers, maximum voting is used.

	\begin{itemize}
		\item \textbf{Adaboost: }Freund and Schapire~\cite{freund19} proposed the adaptive boosting (AdaBoost) algorithm. The idea behind AdaBoost is to use weighted versions of the same training data instead of randomly chosen subsamples, hence same training set could be repeatedly utilized. Due to this reason, training set does not need to be very large.
	\end{itemize}
\end{itemize}

\subsubsection{\textbf{Discussions on the classification frameworks}}
Here, we discuss some salient aspects of the above discussed methods, based on which one can still have some useful comparisons among the above strategies, from various facets.

The cascade system utilizes double the number of binary blocks than the one vs one strategy. However, we note that, in the cascade framework, the first stage which addresses the bulk of the classification problem, works with a small feature space, wherein each classifier works with a only 3 - 6 scalar features, which are selected from the set of class-specific features. Also, in the second stage although high dimensional features are used, the classification involves a very small number of samples (only ATPs). Also, for each sample few binary classification blocks (1 - 3) are actually used, corresponding to the classes for which the ambiguity from the first stage is to be resolved. 

However, because a lot less features are used in the first stage, the classification accuracy for the cascade framework can somewhat relatively suffer. However, we demonstrate that it still yields a reasonably high performance considering the reduction in feature space for a large part of the classification problem 

In addition to its advantage of requiring only N classifiers (equal to number of classes), it also does not employ a greedy feature selection as in the two-stage classification, as there are no individual one-vs-one classification sub-blocks involved. 

However, a disadvantage is that unless there is an oversampling / undersampling carried out, it is apparent that the number of negative samples in each block (from the `rest' class) heavily dominate the number of positive samples. This raises the possibility of inducing a bias while training. On the other hand, as the other structures involve one-vs-one classification blocks, such a bias is less. Indeed, we demonstrate that the drop in the classification accuracy due to such a bias is more serious than that due to considering less number of features in the cascade framework. 

As compared to the SVM based classification strategies, in general, random forest has shown to have many advantages such as an inherent estimating of important features, efficiency on large databases, and handling thousands of input variables, estimating missing data and many more. The main drawback of random forest is over-fitting (in case of noisy dataset)~\cite{breiman}. 

As compared to the SVM based classification strategies, in general, bagging and boosting which fall into an umbrella technique called ensemble learning have shown to have many advantages such as an inherent estimating of important features, efficiency on large databases, and handling thousands of input variables, estimating missing data and many more. Some example of such ensemble frameworks (based on decision tree) are: RF, RUF and Adaboost, considered in the proposed study.
In the experimental section, we will observe that most of the ensemble frameworks perform superior to individual SVM based models, which can be due to various reasons such as: (1) averaging out of biases, (2) reduction in the variance and (3) less chance of over-fitting.

\section{\textbf{Experimentation parameters \& Evaluation metrics}}
In this work, we perform experiments with the data provided with the HEp-2 cell classification contest at ICPR 2014~\cite{Hobson}.  We  use 40 \% data for all classes for training, 30\% for validation and remaining 30\% for testing. The recognition rate is computed over 5 different sets of randomly chosen training samples and average results are reported.

\subsection{\textbf{Parameters}}
In all the experiments, we use $\emph{k}_{b}$ = 5,  $\emph{k}_{o}$ = 10 and $\emph{k}_{i}$ = 5 (parameters used for feature extraction). The preprocessing parameter $\gamma$ and the threshold parameter $T$, for each stage are chosen empirically. 

In first and second stages, for each of the binary SVM classifiers, Gaussian radial basis function kernel with parameters C and gamma is chosen. The value of these corresponding parameters lie in range $10^{3}-10^{9}$ and $0.1-0.001$, respectively.

For RF and RUF, we only decide the number of trees. For the number of features at each tree, we use a
default value which is square root of the total number of features. Number of tress are decided based on graph which is plotted between OOB (out of bag) error  and number of tress. The point where the OOB error reduces negligibly is considered as a good point to fix the value of number of trees. For Adaboost, all default values are used.

\subsection{\textbf{Evaluation metrics}}
We utilize true positive (TP) and false positive (TP) as evaluation measures for fair comparison with other state-of-art methods which also used the same metrics. \\

TP: It is a proportion of positive samples that are correctly identified.

\begin{equation}
TP = (\dfrac{N_{test} -S_{M_{i}}}{N_{test}})\times100 \label{TP}
\end{equation}
Here, $ N_{p} $ denotes the  complete test set and $ S_{M_{i}} $ denotes the set of misclassified samples at the $ i^{th} $ stage for particular class.

False Positive ($FP$) : It is a proportion of negatives samples that are incorrectly classified as positive.
\begin{equation}
FP=1/N\sum^{N}_{i=1} \dfrac{WC(i)}{FS_{i}}
\end{equation}
where $WC(i)$ is the wrongly classified samples of class $ i $ and $FS_{i}$ is the false samples for class $ i$ (for example, for class 1, the samples of other classes (2,3,4,5 and 6) will be the false samples).\\

We note that in our multi-stage strategy (two-stage approaches) for each verification block, each stage can have a different true and false positive rate. Hence, to evaluate performance of proposed first stage two metrics, the metrics of OTP (Overall True Positive) and OFP (Overall false Positive) are used, which in turn depend on TP (True Positive) and FP (False positive) at each stage of hierarchy. We discuss this below in the subsection for the results of the two-stage approaches.

\section{Results \& Discussion}
This section discusses the various results obtained using different classification frameworks. Each framework is evaluated on three feature sets (class-specific, standard texture, combined features). 

\subsection {\textbf{Performance over all samples}}
First, we will discuss the results with the standard protocol, where all samples (positive and intermediate) are considered for training and testing. 

\subsubsection{\textbf{One-vs-One approach}}
As discussed earlier, 
in the one-vs-one strategy, to solve n-class problem, the SVM involves n(n-1)/2 one-versus-one binary models, followed by voting. 
Here, traditionally, the features and parameters used are common across all one-vs-one classifiers. Table~\ref{table5} illustrates the results for one vs. one approach using all feature sets. Following observations can be made through Table~\ref{table5}: (1) Superiority of the class-specific features, both in terms of true and false positives can clearly be noticed over the other feature sets. Hence, it is concluded that the low-dimensional class-specific feature set (128 dimension) which is extracted based on domain knowledge, is adequate to outperform the other considered higher dimensional feature sets. (2) Since class-specific features are thoughtfully designed based on domain knowledge, they yield less false positive as compared to texture features. 

\begin{table}[h!]
	\centering
	\caption{One vs. one approach (figures in \%)}
	\label{table5}
	\medskip
	\setlength\extrarowheight{1pt} 
	\resizebox{13.80cm}{!}{
		
		\begin{tabular}{|c|c|c|c|c|c|c|c|c|c|c|c|c|}
			\hline
			\multirow{3}{*}{} & \multicolumn{4}{c|}{\textbf{Class-specific}}                                                                                                  & \multicolumn{4}{c|}{\textbf{Texture}}                                                                                                         & \multicolumn{4}{c|}{\textbf{Combined}}                                                                                                        \\ \cline{2-13} 
			& \multicolumn{2}{c|}{\textbf{Validation}}                              & \multicolumn{2}{c|}{\textbf{Testing}}                                 & \multicolumn{2}{c|}{\textbf{Validation}}                              & \multicolumn{2}{c|}{\textbf{Testing}}                                 & \multicolumn{2}{c|}{\textbf{Validation}}                              & \multicolumn{2}{c|}{\textbf{Testing}}                                 \\ \cline{2-13} 
			& \multicolumn{1}{l|}{\textbf{TP}} & \multicolumn{1}{l|}{\textbf{FP}} & \multicolumn{1}{l|}{\textbf{TP}} & \multicolumn{1}{l|}{\textbf{FP}} & \multicolumn{1}{l|}{\textbf{TP}} & \multicolumn{1}{l|}{\textbf{FP}} & \multicolumn{1}{l|}{\textbf{TP}} & \multicolumn{1}{l|}{\textbf{FP}} & \multicolumn{1}{l|}{\textbf{TP}} & \multicolumn{1}{l|}{\textbf{FP}} & \multicolumn{1}{l|}{\textbf{TP}} & \multicolumn{1}{l|}{\textbf{FP}} \\ \hline
			\textbf{Class1}   & 98.42                             & 0.14                              & 98.69                             & 0.16                              & 94.55                             & 1.47                              & 94.63                             & 1.52                              & 95.86                             & 0.95                              & 95.56                             & 1.06                              \\ \hline
			\textbf{Class2}   & 97.36                             & 0.48                              & 97.48                             & 0.45                              & 92.72                             & 1.66                              & 93.44                             & 1.63                              & 94.42                             & 1.39                              & 94.73                             & 1.51                              \\ \hline
			\textbf{Class3}   & 98.23                             & 0.8                               & 98.21                             & 0.84                              & 95.33                             & 1.47                              & 95.77                             & 1.51                              & 95.71                             & 1.23                              & 96.00                             & 1.41                              \\ \hline
			\textbf{Class4}   & 99.05                             & 0.12                              & 99.17                             & 0.1                               & 97.91                             & 0.54                              & 97.23                             & 0.54                              & 98.13                             & 0.44                              & 97.35                             & 0.43                              \\ \hline
			\textbf{Class5}   & 97.49                             & 0.68                              & 99.04                             & 0.63                              & 95.92                             & 0.92                              & 95.11                             & 0.68                              & 96.28                             & 0.82                              & 95.08                             & 0.64                              \\ \hline
			\textbf{Class6}   & 96.13                             & 0.18                              & 95.48                             & 0.29                              & 85.25                             & 0.4                               & 86.36                             & 0.5                               & 90.32                             & 0.35                              & 89.68                             & 0.51                              \\ \hline
			\textbf{Avg.}     & 97.78                             & 0.4                               & 97.68                             & 0.4                               & 93.61                             & 1.08                              & 93.76                             & 1.07                              & 95.12                             & 0.86                              & 94.73                             & 0.93                              \\ \hline
			\textbf{F-score}  & \multicolumn{2}{c|}{\textbf{0.9868}}                                  & \multicolumn{2}{c|}{\textbf{0.9863}}                                  & \multicolumn{2}{c|}{\textbf{0.9616}}                                  & \multicolumn{2}{c|}{\textbf{0.9625}}                                  & \multicolumn{2}{c|}{\textbf{0.9707}}                                  & \multicolumn{2}{c|}{\textbf{0.9683}}                                  \\ \hline
	\end{tabular}}
\end{table}

\subsubsection{\textbf{Two-stage approaches}}
\begin{itemize}
	\item \textbf{One-vs-rest approach:} The one-vs-rest approach involves training a single classifier per class, with the samples of that class as positive samples and all other samples as negatives. 
	One important issue with one-vs-rest classification is class imbalance (i.e. having more negative samples than the positive samples). To avoid class imbalance, we maintain an equal number of samples from both classes. As all the classes have a different number of samples, to make sure that each class giving equal contribution, we employ the selection of number of samples as follows:
	
	\begin{equation}
	N_i = \sum_j\alpha N_j \hspace{0.3cm} j \neq i, ~~\alpha = \frac{N_i}{N_{j1}+N_{j2}+N_{j3}+N_{j4}+N_{j5}}
	\end{equation}
	where $N_i$ and $N_j$ are the number of samples in the $i^{\mbox{th}}$ and the $j^{\mbox{th}}$ class, respectively. Here, for each verification block, $N_i$ is considered for the positive class, and $N_j$ for all other classes.
	
	
	As mentioned earlier, (3.4.2), at the testing time, there is possibility that same sample can be classified by more than one block. Hence, we have the notion of the Overall True Positive (OTP) being divided as certain true positives (CTPs), and ambiguous true positives (ATPs).  
	The latter will contribute to the false positives (OFP). After second stage, the resolution of ambiguities, will result in an increment in CTP, and decrement in ATP and hence OFP, while overall OTP is remains the same.
	
	Results for such strategy is illustrated in Table~\ref{table6} and~\ref{table16}. In the first stage, one can notice that while the OTP is quite high, the OFP is also high. Considering the second stage, while the OFP reduces, the OTP is also lower as compared to the one-vs-one strategy. 
	Such a difference can be attributed to the fact that we are using less number of samples to resolve the class imbalance, or the bias induced in case we do not balance the number of samples. 
	
	\begin{table}[h!]
		\centering
		\caption{One vs. rest: class-specific \& texture (figures in \%).}
		\label{table6}
		\medskip
		\setlength\extrarowheight{2.5pt} 
		\resizebox{13.80cm}{!}{
			
			\begin{tabular}{|c|c|c|c|c|c|c|c|c|c|c|c|c|c|c|c|c|c|c|c|c|}
				\hline \hline
				
				\multicolumn{11}{|c|}{\textbf{Class-specific}}   	&\multicolumn{10}{|c|}{\textbf{Texture}}                                                                                                                              \\ \hline \hline
				\multirow{2}{*}{\textbf{}} & \multicolumn{5}{c|}{\textbf{Validation}}                                                                                                & \multicolumn{5}{c|}{\textbf{Testing}}    & \multicolumn{5}{c|}{\textbf{Validation}}                                                                                                & \multicolumn{5}{c|}{\textbf{Testing}}                                                                                                 \\ \cline{2-21} 
				
				\multirow{3}{*}{\textbf{}} & \multicolumn{3}{c|}{\textbf{First stage}}                                                                                                & \multicolumn{2}{c|}{\textbf{Second stage}}    & \multicolumn{3}{c|}{\textbf{First stage}}                                                                                                & \multicolumn{2}{c|}{\textbf{Second stage}}     & \multicolumn{3}{c|}{\textbf{First stage}}                                                                                                & \multicolumn{2}{c|}{\textbf{Second stage}}    & \multicolumn{3}{c|}{\textbf{First stage}}                                                                                                & \multicolumn{2}{c|}{\textbf{Second stage}}                                                                                              \\ \cline{2-21} 
				\multirow{3}{*}{\textbf{}} & \multicolumn{2}{c|}{\textbf{OTP}}                                                                                                & \multicolumn{1}{c|}{\textbf{OFP}}    & \multicolumn{1}{c|}{\textbf{OTP}}                                                                                                & \multicolumn{1}{c|}{\textbf{OFP}}     &
				\multicolumn{2}{c|}{\textbf{OTP}}                                                                                                & \multicolumn{1}{c|}{\textbf{OFP}}    & \multicolumn{1}{c|}{\textbf{OTP}}                                                                                                & \multicolumn{1}{c|}{\textbf{OFP}}     & \multicolumn{2}{c|}{\textbf{OTP}}                                                                                                & \multicolumn{1}{c|}{\textbf{OFP}}    & \multicolumn{1}{c|}{\textbf{OTP}}                                                                                                & \multicolumn{1}{c|}{\textbf{OFP}}     &
				\multicolumn{2}{c|}{\textbf{OTP}}                                                                                                & \multicolumn{1}{c|}{\textbf{OFP}}    & \multicolumn{1}{c|}{\textbf{OTP}}                                                                                                & \multicolumn{1}{c|}{\textbf{OFP}}                                                                                              \\ \cline{2-21}

				& \textbf{CTP}      & \textbf{ATP}  & \textbf{}   & \textbf{} & \textbf{}  & \textbf{CTP}      & \textbf{ATP}   & \textbf{}   & \textbf{} & \textbf{} 	& \textbf{CTP}      & \textbf{ATP}   & \textbf{}  & \textbf{} & \textbf{}  & \textbf{CTP}      & \textbf{ATP}  & \textbf{}   & \textbf{} & \textbf{}\\ \hline	
				
				\textbf{Class1}     & 90.93             & 7.8       & 0.59      & 97.59        & 0.23            
				& 91.12             & 7.7       &0.59      & 97.59        & 0.31         
				
				& 81.55             & 15.42     &3.98       & 93.71        & 2.88              & 80.21             & 16.17      &4.23       & 92.83        & 2.95  \\ \hline	
				
				\textbf{Class2}                 & 92.67             & 4.80         & 3.82   & 97.17        & 1.18           
				& 93.85             & 4.14        &3.65    & 97.36        & 1.19  
				
				& 80.77             & 14.67        &4.97      & 88.66        & 1.91  
				& 80.94             & 15.34     &5.41        & 89.69        & 1.93      \\ \hline	
				
				\textbf{Class3}                 & 89.13             & 8.39   &2.58         & 95.30        & 0.98                
				& 89.92        & 7.43      &2.6       & 95.43        & 1.01    
				
				& 75.81            & 20.77      &4.36      & 93.96        & 2.22         
				& 75.97             & 20.58      &4.11      & 93.94        & 2.01        \\ \hline
				
				\textbf{Class4}                & 95.57             & 2.94      &0.71       & 97.54        & 0.18             
				& 95.62            & 2.77        &0.77     & 97.44        & 0.21    
				& 92.43             & 5.96      &1.64       & 97.12        & 0.52                 & 91.83             & 6.17      &1.92       & 96.40        & 0.54       \\ \hline
				
				\textbf{Class5}                 & 88.36             & 9.27       &2.65     & 95.31        & 1.06              
				& 87.42             & 10.0      &2.35       & 94.87       & 1.05     
				&76.40             & 21.54       &2.20      & 85.16         & 0.91                 & 76.31              & 20.69     &2.18        & 84.73        & 0.87    \\ \hline

				\textbf{Class6}                  & 84.14             & 14.65      &1.01       & 96.03        & 0.53              
				& 85.25             & 12.99       &1.23     & 96.31        & 0.67        
				& 62.11            & 31.52       &6.71      & 84.33        & 2.26                & 63.87             & 30.04      &6.77      & 86.35        & 2.46   \\ \hline
				
				\textbf{Avg.}                   & 90.13             & 7.89     &1.89        & 96.49        & 0.69                
				& 90.53            & 7.50         &1.86    & 96.52        & 0.74    
				& 78.17            & 18.31    &3.97        & 90.49        & 1.79                 & 78.19             & 18.17      &4.10      & 90.60       & 1.79    \\ \hline

				\textbf{F-score}  & \textbf{}  & \textbf{}  & \textbf{}& \multicolumn{2}{c|}{\textbf{0.9787}}        & \textbf{}  & \textbf{}  & \textbf{} & \multicolumn{2}{c|}{\textbf{0.9786}}        & \textbf{}   & \textbf{}  & \textbf{} & \multicolumn{2}{c|}{\textbf{0.9412}}     & \textbf{}  & \textbf{}   & \textbf{} & \multicolumn{2}{c|}{\textbf{0.9418}}   \\ \hline 
		\end{tabular}}
	\end{table}

	\begin{table}[h!]
		\centering
		\caption{One vs. rest: Combined (figures in \%).}
		\label{table16} 
		\medskip
		\setlength\extrarowheight{2.5pt} 
		\resizebox{9.64cm}{!}{
			
			\begin{tabular}{|c|c|c|c|c|c|c|c|c|c|c|}
				\hline \hline
				\multicolumn{11}{|c|}{\textbf{Combined}}                                                                                                                           \\ \hline \hline	
				\multirow{2}{*}{\textbf{}} & \multicolumn{5}{c|}{\textbf{Validation}}                                                                                                & \multicolumn{5}{c|}{\textbf{Testing}}                                                                                                   \\ \cline{2-11}
				\multirow{3}{*}{\textbf{}} & \multicolumn{3}{c|}{\textbf{First stage}}                                                                                                & \multicolumn{2}{c|}{\textbf{Second stage}}    & \multicolumn{3}{c|}{\textbf{First stage}}                                                                                                & \multicolumn{2}{c|}{\textbf{Second stage}}                                                                                                \\ \cline{2-11} 
				\multirow{3}{*}{\textbf{}} & \multicolumn{2}{c|}{\textbf{OTP}}                                                                                                & \multicolumn{1}{c|}{\textbf{OFP}}    & \multicolumn{1}{c|}{\textbf{OTP}}                                                                                                & \multicolumn{1}{c|}{\textbf{OFP}}     &
				\multicolumn{2}{c|}{\textbf{OTP}}                                                                                                & \multicolumn{1}{c|}{\textbf{OFP}}    & \multicolumn{1}{c|}{\textbf{OTP}}                                                                                                & \multicolumn{1}{c|}{\textbf{OFP}}                                                                                            \\ \cline{2-11} 
				& \textbf{CTP}      & \textbf{ATP}  & \textbf{}   & \textbf{} & \textbf{}  & \textbf{CTP}      & \textbf{ATP}   & \textbf{}   & \textbf{} & \textbf{}  \\ \hline
				
				\textbf{Class1}                  & 76.89            & 20.74        &3.08     & 93.36        & 2.07                & 76.84             & 20.48    &3.40        & 92.29        & 2.04         \\ \hline
				\textbf{Class2}                   & 83.15             & 13.26      &6.31      & 90.81        & 1.93                & 83.22             & 13.92      &6.23      & 91.43        & 2.04         \\ \hline
				\textbf{Class3}                    & 82.84             & 13.76    &4.22        & 94.30        & 1.79                 & 83.15             & 13.52     &4.06       & 94.43        & 1.64         \\ \hline
				\textbf{Class4}                    & 92.40             & 6.15      &1.49      & 96.98       & 0.33                & 92.34             & 6.00          &1.71   & 96.71       & 0.28         \\ \hline
				\textbf{Class5}                    & 81.99            & 15.98     &2.13        & 95.10        & 1.02                & 80.27             & 16.80     &2.04        & 94.23        & 0.94         \\ \hline
				\textbf{Class6}                  & 63.13             & 33.64      &4.65       & 85.80        & 1.03                & 66.26             & 30.68        &4.52     & 87.37        & 1.33         \\ \hline
				\textbf{Avg.}                     & 80.06            & 17.25      &3.64        & 92.72        & 1.36                & 80.34             &16.90        &3.66     & 92.75        & 1.38         \\ \hline
				\textbf{F-score}        & \textbf{}       & \textbf{}  & \textbf{} & \multicolumn{2}{c|}{\textbf{0.9555}}       & \textbf{} & \textbf{} & \textbf{} & \multicolumn{2}{c|}{\textbf{0.9555}}        \\ \hline
		\end{tabular}}
	\end{table}

	\item \textbf{Hierarchical approach:}
	In the hierarchical strategy, for each verification block, each stage can have a different true and false positive rate. Hence, to evaluate performance of proposed first stage two metrics, OTP (Overall True Positive) and OFP (Overall false Positive) are used, which in turn depend on TP (True Positive) and FP (False positive) at each stage of hierarchy. Again, the true positive samples at the end of the first stage (OTP) are divided as certain true positives (CTPs) and ambiguous true positives (ATPs). 
	Below we describe the terms which are required to evaluate the overall performance.

	 $N_{p}$, $S_{M_{i}} $ and $ S_{C_{i}} $ denote  the complete test set, set of misclassified samples at the $ i^{th} $ stage, and set of correctly classified samples at the $ i^{th} $ stage for each class respectively.

	OTP: It is a fraction of samples which are not rejected at the end of all stages. In most cases, large portions of falsely-rejected positive samples may overlap between stages. Hence this enumeration can be given as follows.
	\begin{equation}
	OTP = (\dfrac{N_{test} -M_{s}}{N_{test}})\times100, M_{s} =\mid\underset{i}\cup S_{M_{i}}\mid \nonumber
	\end{equation}
	The symbols $\cup$, $\mid\mid $, $ M_{s} $ denotes the union operator, set cardinality, and number of miss-classified samples at the end of all stages,, respectively.
	
	OFP: It is a fraction of samples which are common across  all the stages. Only these fraction of samples allowed to be recognized as a particular(true) class.
	\begin{equation}
	OFP =\frac{W_{s}}{N_{test}} \times 100, W_{s} =\mid\underset{i}\cap S_{C_{i}}\mid  \nonumber
	\end{equation}
	The symbols, $\cap$, $W_{s}$ denotes the intersection operator and number of classified samples at the end of all stages, respectively.

	\begin{table}[h!]
		\centering
		\caption{Cascade framework: class-specific(1)(2) \& class-specific-texture(1)(2) (figures in \%).}
		\label{table3} 
		\medskip
		\setlength\extrarowheight{2.5pt} 
		\resizebox{13.80cm}{!}{
			
			\begin{tabular}{|c|c|c|c|c|c|c|c|c|c|c|c|c|c|c|c|c|c|c|c|c|}
				\hline \hline
				
				\multicolumn{11}{|c|}{\textbf{Class-specific-low-dimensional(1), Class-specific-high dimensional(2)}}   	&\multicolumn{10}{|c|}{\textbf{Class-specific-low-dimensional(1)-Texture(2)}}                                                                                                                              \\ \hline \hline
				\multirow{2}{*}{\textbf{}} & \multicolumn{5}{c|}{\textbf{Validation}}                                                                                                & \multicolumn{5}{c|}{\textbf{Testing}}    & \multicolumn{5}{c|}{\textbf{Validation}}                                                                                                & \multicolumn{5}{c|}{\textbf{Testing}}                                                                                                 \\ \cline{2-21} 
				
				\multirow{3}{*}{\textbf{}} & \multicolumn{3}{c|}{\textbf{First stage}}                                                                                                & \multicolumn{2}{c|}{\textbf{Second stage}}    & \multicolumn{3}{c|}{\textbf{First stage}}                                                                                                & \multicolumn{2}{c|}{\textbf{Second stage}}     & \multicolumn{3}{c|}{\textbf{First stage}}                                                                                                & \multicolumn{2}{c|}{\textbf{Second stage}}    & \multicolumn{3}{c|}{\textbf{First stage}}                                                                                                & \multicolumn{2}{c|}{\textbf{Second stage}}                                                                                              \\ \cline{2-21} 
				\multirow{3}{*}{\textbf{}} & \multicolumn{2}{c|}{\textbf{OTP}}                                                                                                & \multicolumn{1}{c|}{\textbf{OFP}}    & \multicolumn{1}{c|}{\textbf{OTP}}                                                                                                & \multicolumn{1}{c|}{\textbf{OFP}}     &
				\multicolumn{2}{c|}{\textbf{OTP}}                                                                                                & \multicolumn{1}{c|}{\textbf{OFP}}    & \multicolumn{1}{c|}{\textbf{OTP}}                                                                                                & \multicolumn{1}{c|}{\textbf{OFP}}     & \multicolumn{2}{c|}{\textbf{OTP}}                                                                                                & \multicolumn{1}{c|}{\textbf{OFP}}    & \multicolumn{1}{c|}{\textbf{OTP}}                                                                                                & \multicolumn{1}{c|}{\textbf{OFP}}     &
				\multicolumn{2}{c|}{\textbf{OTP}}                                                                                                & \multicolumn{1}{c|}{\textbf{OFP}}    & \multicolumn{1}{c|}{\textbf{OTP}}                                                                                                & \multicolumn{1}{c|}{\textbf{OFP}}                                                                                              \\ \cline{2-21}

				& \textbf{CTP}      & \textbf{ATP}  & \textbf{}   & \textbf{} & \textbf{}  & \textbf{CTP}      & \textbf{ATP}   & \textbf{}   & \textbf{} & \textbf{} 	& \textbf{CTP}      & \textbf{ATP}   & \textbf{}  & \textbf{} & \textbf{}  & \textbf{CTP}      & \textbf{ATP}  & \textbf{}   & \textbf{} & \textbf{}\\ \hline	
				
				\textbf{Class1}    & 18.26        & 78.26    & 20.55     &96.31        & 1.61     
				& 19.49        & 76.66     &  20.86            & 95.82             & 1.37         
				& 18.26        & 78.26    & 20.55       & 92.24             & 2.51     
				& 19.49        & 76.66     &  20.86             & 92.29             & 2.29               \\ \hline	
				
				\textbf{Class2}     & 7.61       & 90.18 &48.05      & 96.32             & 2.76            
				
				& 3.51        & 93.78  &47.93  & 96.32            & 3.26                
				
				& 7.61       & 90.18 &48.05      & 92.93            & 4.1                   
				
				& 3.51        & 93.78  &47.93   & 92.54             & 4.53                  \\ \hline	
				
				\textbf{Class3}        & 16.84        & 77.72          &16.12                & 93.99            & 1.72            
				& 18.41        & 75.44  &16.41 & 93.33             & 1.98
				
				& 16.84        & 77.72          &16.12  & 92.42             & 2.07    
				& 18.41        & 75.44  &16.41 & 91.76             & 2.41                    \\ \hline
				
				\textbf{Class4}     & 66.69        & 31.85 &20.33                        & 98.12             & 1.36            
				& 66.63        & 31.13  &19.61 & 97.30            & 1.48              
				& 66.69        & 31.85 &20.33     & 97.68            & 1.52        
				& 66.63        & 31.13  &19.61                 & 96.23             & 1.76                   \\ \hline
				
				\textbf{Class5}         & 58.67        & 36.31   &12.07                  & 94.16             & 1.06             
				& 57.83        & 35.93 &11.76    & 92.67             & 1.12             
				& 58.67        & 36.31   &12.07   & 93.5             & 1.24     
				& 57.83        & 35.93 &11.76                       & 92.33              & 1.18                 \\ \hline

				\textbf{Class6}     & 70.23        & 23.4     & 0.19                     & 88.29             & 0.06            
				& 70.41       & 19.54    &0.2 & 89.21            & 0.08               
				& 70.23        & 23.4     & 0.19   & 86.17             & 0.06           
				& 70.41       & 19.54    &0.2                  & 86.82             & 0.088              \\ \hline
				
				\textbf{Avg.}         & 39.71        & 56.28 &19.55                   & 94.54             & 1.43            
				& 39.38        & 55.41 &19.46  & 94.14            & 1.55               
				& 39.71        & 56.28 &19.55     & 92.49             & 1.91       
				& 39.38        & 55.41 &19.46                 & 91.99            & 2.043                \\ \hline

				\textbf{F-score}  & \textbf{}  & \textbf{} & \textbf{}& \multicolumn{2}{c|}{\textbf{0.9668}}         & \textbf{} & \textbf{}  & \textbf{}& \multicolumn{2}{c|}{\textbf{0.9644}}         & \textbf{}  & \textbf{} & \textbf{} & \multicolumn{2}{c|}{\textbf{0.9576}}      & \textbf{}   & \textbf{} & \textbf{} & \multicolumn{2}{c|}{\textbf{0.9545}}   \\ \hline 
		\end{tabular}}
	\end{table}

	\begin{table}[h!]
		\centering
		\caption{Cascade framework: class-specific-combined (figures in \%).}
		\label{table11} 
		\medskip
		\setlength\extrarowheight{2.5pt} 
		\resizebox{9.64cm}{!}{
			
			\begin{tabular}{|c|c|c|c|c|c|c|c|c|c|c|}
				\hline \hline
				\multicolumn{11}{|c|}{\textbf{Class-specific-low-dimensional(1)-Combined(2) }}                                                                                                                           \\ \hline \hline	
				\multirow{2}{*}{\textbf{}} & \multicolumn{5}{c|}{\textbf{Validation}}                                                                                                & \multicolumn{5}{c|}{\textbf{Testing}}                                                                                                   \\ \cline{2-11}
				\multirow{3}{*}{\textbf{}} & \multicolumn{3}{c|}{\textbf{First stage}}                                                                                                & \multicolumn{2}{c|}{\textbf{Second stage}}    & \multicolumn{3}{c|}{\textbf{First stage}}                                                                                                & \multicolumn{2}{c|}{\textbf{Second stage}}                                                                                                \\ \cline{2-11} 
				\multirow{3}{*}{\textbf{}} & \multicolumn{2}{c|}{\textbf{OTP}}                                                                                                & \multicolumn{1}{c|}{\textbf{OFP}}    & \multicolumn{1}{c|}{\textbf{OTP}}                                                                                                & \multicolumn{1}{c|}{\textbf{OFP}}     &
				\multicolumn{2}{c|}{\textbf{OTP}}                                                                                                & \multicolumn{1}{c|}{\textbf{OFP}}    & \multicolumn{1}{c|}{\textbf{OTP}}                                                                                                & \multicolumn{1}{c|}{\textbf{OFP}}                                                                                            \\ \cline{2-11} 
				& \textbf{CTP}      & \textbf{ATP}  & \textbf{}   & \textbf{} & \textbf{}  & \textbf{CTP}      & \textbf{ATP}   & \textbf{}   & \textbf{} & \textbf{}  \\ \hline
				
				\textbf{Class1}      & 18.26        & 78.26 &20.55                          & 93.53             & 2.21            
				& 19.49        & 76.66 &20.86    & 93.18             & 2.07                    \\ \hline
				\textbf{Class2}        & 7.61        & 90.18 &48.05                           & 94.35             & 3.71            
				& 3.51        & 93.78   &47.93  & 93.48             & 4.36                   \\ \hline
				\textbf{Class3}      & 16.84        & 77.72           &16.12                   & 92.43             & 2.09             
				& 18.41       & 75.44 &16.41   & 91.95            & 2.39                   \\ \hline
				\textbf{Class4}        & 66.69        & 31.85 &20.33                             & 97.62            & 1.49           
				& 66.63       & 31.13 &19.61   & 96.26             & 1.71                    \\ \hline
				\textbf{Class5}      & 58.67        & 36.31 &12.07                              & 93.84             & 1.26             
				& 57.83        & 35.93 &11.76 & 92.01             & 1.16                      \\ \hline
				\textbf{Class6}    & 70.23        & 23.40 &0.19                             & 86.91             & 0.06             
				& 70.41        & 19.54 &0.2  	& 87.28             & 0.09                     \\ \hline
				\textbf{Avg.}        & 39.71        & 56.28 & 19.55                             & 93.11            & 1.8             
				& 39.85        & 55.41 &19.46   & 92.36             & 1.96                     \\ \hline
				\textbf{F-score}        & \textbf{}    & \textbf{}  & \textbf{} & \multicolumn{2}{c|}{\textbf{0.9606}}     & \textbf{}   & \textbf{}  & \textbf{}& \multicolumn{2}{c|}{\textbf{0.9569}}         \\ \hline
		\end{tabular}}
	\end{table}	
	
	We provide the results in Table~\ref{table3} and \ref{table11}. In all the tables, notation (1) \& (2) specify the feature set used at first and second stage. Following can be observed from the Table~\ref{table3} and \ref{table11}: (1) A higher classification performance is obtained using class-specific features in both the stages. The low dimensional class-specific features utilized at first stage yield a high true positive rate but somewhat high false positive. This indicates that while test sample is identified by its corresponding block, in a high number of cases, in some cases, this sample may also be identified by a block of another class. However, we note that such an ambiguity is between only 2 classes for 97\% cases, which supports the point that just the first stage can also be used to reduce ambiguity to few classes. 
	(2) There is a small decrease in the OTP (or CTP after the second stage), than that from the first stage. However, this is much less significant as compared to the drop in the OFP. From a systems perspective, we believe that, achieving a high OTP (even with ambiguities) is more important, as such ambiguities can be reduced in the next stage.
	(3) Importantly, we note that while the performance is lower than (but close to) the one-vs-one classifier, it is higher than the one-vs-rest case. A computational advantage here is that a small number of features are used in most binary classification blocks. Thus, we believe that it is interesting to consider such differently designed frameworks.
	
	
	\item \textbf{Hierarchy with common features:}
	Results for hierarchical structure where same features but different parameter settings  are
	employed at each step of verification block, are shown in table 9 and table 10. 
	It is apparent from the tables that this strategy produces better accuracy than the other two-stage
	approaches like one-vs-rest and two-stage hierarchical cascade framework, while it yields somewhat similar
	accuracy to the one-vs-one approach. The reason for such behavior could be: (1) Feature vector which includes characteristics of all the classes is utilized at all the stages with different parameter setting. While the one-vs-rest also uses the same features  for each verification block, the flexibility of different parameters at each sub-block in this case of hierarchical structure helps the classification. 
	Also, unlike one-vs-rest approach, each sub-block considers only a one-vs-one problem, which is similar to the case of one-vs-one.  (2) The improvement in accuracy over two-stage cascade framework relates to the fact that more features are involved in this case. Perhaps, this enables better learning of characteristics of the classes which was missed in when less number of features per sub-block are used in the earlier case.  Hence, we note that, involvement of verification blocks at first stage itself plays an important role along with features.
	
	
	\begin{table}[h!]
		\centering
		\caption{Hierarchy with common feature: Class-specific \& texture features (figures in \%).}
		\label{table13} 
		\medskip
		\setlength\extrarowheight{2.5pt} 
		\resizebox{13.80cm}{!}{
			
			\begin{tabular}{|c|c|c|c|c|c|c|c|c|c|c|c|c|c|c|c|c|c|c|c|c|}
				\hline \hline
				
				\multicolumn{11}{|c|}{\textbf{Class-specific}}   	&\multicolumn{10}{|c|}{\textbf{Texture}}                                                                                                                              \\ \hline \hline
				\multirow{2}{*}{\textbf{}} & \multicolumn{5}{c|}{\textbf{Validation}}                                                                                                & \multicolumn{5}{c|}{\textbf{Testing}}    & \multicolumn{5}{c|}{\textbf{Validation}}                                                                                                & \multicolumn{5}{c|}{\textbf{Testing}}                                                                                                 \\ \cline{2-21} 
				
				\multirow{3}{*}{\textbf{}} & \multicolumn{3}{c|}{\textbf{First stage}}                                                                                                & \multicolumn{2}{c|}{\textbf{Second stage}}    & \multicolumn{3}{c|}{\textbf{First stage}}                                                                                                & \multicolumn{2}{c|}{\textbf{Second stage}}     & \multicolumn{3}{c|}{\textbf{First stage}}                                                                                                & \multicolumn{2}{c|}{\textbf{Second stage}}    & \multicolumn{3}{c|}{\textbf{First stage}}                                                                                                & \multicolumn{2}{c|}{\textbf{Second stage}}                                                                                              \\ \cline{2-21} 
				\multirow{3}{*}{\textbf{}} & \multicolumn{2}{c|}{\textbf{OTP}}                                                                                                & \multicolumn{1}{c|}{\textbf{OFP}}    & \multicolumn{1}{c|}{\textbf{OTP}}                                                                                                & \multicolumn{1}{c|}{\textbf{OFP}}     &
				\multicolumn{2}{c|}{\textbf{OTP}}                                                                                                & \multicolumn{1}{c|}{\textbf{OFP}}    & \multicolumn{1}{c|}{\textbf{OTP}}                                                                                                & \multicolumn{1}{c|}{\textbf{OFP}}     & \multicolumn{2}{c|}{\textbf{OTP}}                                                                                                & \multicolumn{1}{c|}{\textbf{OFP}}    & \multicolumn{1}{c|}{\textbf{OTP}}                                                                                                & \multicolumn{1}{c|}{\textbf{OFP}}     &
				\multicolumn{2}{c|}{\textbf{OTP}}                                                                                                & \multicolumn{1}{c|}{\textbf{OFP}}    & \multicolumn{1}{c|}{\textbf{OTP}}                                                                                                & \multicolumn{1}{c|}{\textbf{OFP}}                                                                                              \\ \cline{2-21}

				& \textbf{CTP}      & \textbf{ATP}  & \textbf{}   & \textbf{} & \textbf{}  & \textbf{CTP}      & \textbf{ATP}   & \textbf{}   & \textbf{} & \textbf{} 	& \textbf{CTP}      & \textbf{ATP}   & \textbf{}  & \textbf{} & \textbf{}  & \textbf{CTP}      & \textbf{ATP}  & \textbf{}   & \textbf{} & \textbf{}\\ \hline	
				
				\textbf{Class1}     & 98.07             & 0.64          &2.41   & 98.15        & 0.13           
				& 98.23           & 0.74      &2.24       & 98.28       & 0.14        
				& 35.66              & 57.54       &1.45     & 93.18        & .74             
				& 36.71             & 56.78        &1.51     & 93.36        & 0.75  \\ \hline	
				
				\textbf{Class2}                 & 95.83             & 0.89       &0.60      & 95.90        & 0.30            
				& 96.42            & 0.89     &0.50        & 96.44        & 0.32    
				& 92.65             & 2.94     &14.48         & 95.40        & 1.64            
				& 93.22            & 3.01     &14.82        & 95.92        & 1.65      \\ \hline	
				
				\textbf{Class3}                 & 97.04             & 0.66        &0.73     & 97.66        & 0.69               
				& 97.61             & 0.35        &0.73     & 97.97       & 0.71    
				& 93.32             & 1.66      &2.04       & 93.78        & 1.49               
				& 94.02             & 1.35      &2.05      & 94.35        & 1.44        \\ \hline
				
				\textbf{Class4}                & 98.34             & 0.09        &0.12    & 98.34        & 0.10             
				& 98.46             & 0.26       &0.17      & 98.49        & 0.10   
				& 96.73            & 0.41        &0.45     & 97.15       & 0.45                
				& 95.69             & 0.75     &0.57        & 96.42        & 0.57       \\ \hline
				
				\textbf{Class5}                & 95.56             & 1.8   &0.70          & 97.16        & 0.74              
				& 94.66             & 2.01    &0.71         &96.41        & 0.70     
				& 92.05             & 4.08    &0.87         & 94.41         & 0.79                & 91.55              & 3.95      &0.66      & 94.05       & 0.64    \\ \hline

				\textbf{Class6}                  & 96.49             & 0.18      &0.18       & 96.68        & 0.00              
				& 94.28             & 0.82        &0.25    & 95.02        & 0.25       
				& 83.04            & 0.82      &0.34       & 83.13        & .34               
				& 83.13            & 0.73      &0.46      & 83.13        & 0.42   \\ \hline
				
				\textbf{Avg.}                   & 95.91            & 2.49   &0.79          & 97.31        & 0.29               
				& 95.03             & 2.42       &0.76      & 97.11        & 0.37    
				& 82.24            & 0.82       &3.27      & 92.84        & 0.91                 & 82.39             & 11.08      &3.71      & 92.87        & 0.91    \\ \hline

				\textbf{F-score}  & \textbf{}  & \textbf{} & \textbf{} & \multicolumn{2}{c|}{\textbf{0.9849}}         & \textbf{} & \textbf{}  & \textbf{} & \multicolumn{2}{c|}{\textbf{0.9835}}        & \textbf{}  & \textbf{} & \textbf{}& \multicolumn{2}{c|}{\textbf{0.9583}}       & \textbf{}   & \textbf{}  & \textbf{}&   \multicolumn{2}{c|}{\textbf{0.9585}}   \\ \hline 
		\end{tabular}}
	\end{table}
	
	\begin{table}[h!]
		\centering
		\caption{Hierarchy with common feature: combined features (figures in \%).}
		\label{table12} 
		\medskip
		\setlength\extrarowheight{2.5pt} 
		\resizebox{9.64cm}{!}{
			
			\begin{tabular}{|c|c|c|c|c|c|c|c|c|c|c|}
				\hline \hline
				\multicolumn{11}{|c|}{\textbf{Combined}}                                                                                                                           \\ \hline \hline	
				\multirow{2}{*}{\textbf{}} & \multicolumn{5}{c|}{\textbf{Validation}}                                                                                                & \multicolumn{5}{c|}{\textbf{Testing}}                                                                                                   \\ \cline{2-11}
				\multirow{3}{*}{\textbf{}} & \multicolumn{3}{c|}{\textbf{First stage}}                                                                                                & \multicolumn{2}{c|}{\textbf{Second stage}}    & \multicolumn{3}{c|}{\textbf{First stage}}                                                                                                & \multicolumn{2}{c|}{\textbf{Second stage}}                                                                                                \\ \cline{2-11} 
				\multirow{3}{*}{\textbf{}} & \multicolumn{2}{c|}{\textbf{OTP}}                                                                                                & \multicolumn{1}{c|}{\textbf{OFP}}    & \multicolumn{1}{c|}{\textbf{OTP}}                                                                                                & \multicolumn{1}{c|}{\textbf{OFP}}     &
				\multicolumn{2}{c|}{\textbf{OTP}}                                                                                                & \multicolumn{1}{c|}{\textbf{OFP}}    & \multicolumn{1}{c|}{\textbf{OTP}}                                                                                                & \multicolumn{1}{c|}{\textbf{OFP}}                                                                                            \\ \cline{2-11} 
				& \textbf{CTP}      & \textbf{ATP}  & \textbf{}   & \textbf{} & \textbf{}  & \textbf{CTP}      & \textbf{ATP}   & \textbf{}   & \textbf{} & \textbf{}  \\ \hline
				\textbf{Class1}                  & 90.26             & 0.74        &0.83     & 91.01        & 0.76                & 90.02             &0.56    &0.56        & 90.58        & 0.82         \\ \hline
				\textbf{Class2}                   & 92.81             & 0.58       &1.30      & 93.40        & 1.08                & 93.36             & 0.68       &0.58      & 94.04        & 1.71        \\ \hline
				\textbf{Class3}                   & 94.42            & 0.97      &0.98        & 94.96        & 0.90                 & 94.87            & 0.94       &0.97      & 95.28        & 1.00         \\ \hline
				\textbf{Class4}                    & 96.52            & 0.31    &0.36         & 96.64        & 0.32                 & 96.11            & 0.51       &0.31      & 96.20        & 0.33         \\ \hline
				\textbf{Class5}                    & 95.10             & 0.63    &0.90        & 95.43       & 0.80                & 93.84             & 0.63       &0.63    & 94.11       & 0.66        \\ \hline
				\textbf{Class6}                  & 89.40             & 0.69      &0.35      & 86.58        & 0.31                & 88.57             & 1.10       &0.69     & 88.84        &0.45         \\ \hline
				\textbf{Avg.}                     & 93.08             & 0.69     &0.69         & 93.50       & 0.70                & 92.79             & 0.74      &0.74      & 93.18        & 0.74         \\ \hline
				\textbf{F-score}        & \textbf{}    & \textbf{}  & \textbf{} & \multicolumn{2}{c|}{\textbf{0.9629}}     & \textbf{}    & \textbf{}  & \textbf{} & \multicolumn{2}{c|}{\textbf{0.961}}        \\ \hline
		\end{tabular}}
	\end{table}	

\end{itemize}

\subsubsection{Ensemble frameworks}
This subsection discusses the results obtained using ensemble frameworks. The motivation behind adding this framework for comparison is that, all the above considered frameworks are essentially an ensemble where base classifier (SVM) are arranged in various ways. In the same spirit, we also consider frameworks where decision tree is used as base learner, different from above framework where SVM is used as base learner. 

Table~\ref{table7} illustrates the testing results of various decision tree based ensemble frameworks considering class-specific feature set. For all frameworks, best decision parameters are decided using validation dataset. For example, to decide the value of RF parameters such as number of trees, and number of features used at each tree, the OOB (out of bag) error rate is considered. In RF, we only decide the number of trees. For the number of features at each tree, we use a default value which is square root of the total number of features. 

\begin{table}[h!]
	\centering
	\caption{Performance of existing ensemble frameworks (figures in \%).}
	\label{table7}
	\medskip
	\setlength\extrarowheight{1pt} 
	\resizebox{13.80cm}{!}{
		\begin{tabular}{|c|c|c|c|c|c|c|c|c|c|c|c|c|}
			\hline
			\multirow{3}{*}{\textbf{}} & \multicolumn{4}{c|}{\textbf{Class-specific: RF}}                                     & \multicolumn{4}{c|}{\textbf{Class-specific: RUF}}                                            & \multicolumn{4}{c|}{\textbf{Class-specific: Adaboost}}                                           \\ \cline{2-13} 
			& \multicolumn{2}{c|}{\textbf{Validation}} & \multicolumn{2}{c|}{\textbf{Testing}} & \multicolumn{2}{c|}{\textbf{Validation}} & \multicolumn{2}{c|}{\textbf{Testing}} & \multicolumn{2}{c|}{\textbf{Validation}} & \multicolumn{2}{c|}{\textbf{Testing}} \\ \cline{2-13} 
			& \textbf{OTP}        & \textbf{OFP}       & \textbf{OTP}      & \textbf{OFP}      & \textbf{OTP}        & \textbf{OFP}       & \textbf{OTP}      & \textbf{OFP}      & \textbf{OTP}        & \textbf{OFP}       & \textbf{OTP}      & \textbf{OFP}      \\ \hline
			\textbf{Class1}            & 98.64               & 0.30               & 98.29             & 0.38              & 98.15               & 0.41               & 98.34             & 0.37              & 99.47               & 0.12               & 98.85             & 0.26              \\ \hline
			\textbf{Class2}            & 97.46               & 0.86               & 97.18             & 0.97               & 97.12               & 0.98               & 98.37             & 0.55              & 99.28               & 0.24               & 98.78             & 0.42              \\ \hline
			\textbf{Class3}            & 95.72               & 1.03               & 96.54             & 0.82             & 97.93               & 0.48               & 97.71             & 0.54              & 99.28               & 0.1               & 98.28             & 0.41              \\ \hline
			\textbf{Class4}            & 99.32               & 0.17               & 98.91             & 0.28              & 99.17               & 0.20                & 98.93             & 0.26              & 99.0               & 0.02               & 99.25             & 0.19              \\ \hline
			\textbf{Class5}            & 95.29               & 0.92               & 95.26             & 0.92              & 96.28               & 0.72               & 995.80             & 0.81             & 99.11              & 0.11               & 97.07             & 0.57              \\ \hline
			\textbf{Class6}            & 94.68               & 0.3                & 93.64             & 0.36              & 96.01               & 0.22               & 95.39             & 0.24              & 99.19               & .03               & 96.87             & 0.18              \\ \hline
			\textbf{Avg.}              & 96.85               & 0.60               & 96.64             & 0.62              & 97.44              & 0.50               & 97.42             & 0.46              & 99.22               & 0.10              & 98.18              & 0.34             \\ \hline
			\textbf{F-score}           & \multicolumn{2}{c|}{\textbf{0.981}}     & \multicolumn{2}{c|}{\textbf{0.9798}}  & \multicolumn{2}{c|}{\textbf{0.9845}}     & \multicolumn{2}{c|}{\textbf{0.9840}}  & \multicolumn{2}{c|}{\textbf{0.9956}}     & \multicolumn{2}{c|}{\textbf{0.9891}}   \\ \hline
	\end{tabular}}
\end{table}

Following observations are made from Table~\ref{table7}: (1) Boosting outperforms bagging and plain classifier. This is also shown in the paper~\cite{quinlan} where the author makes comparisons between bagging, boosting and C4.5 over two dozen datasets, and shows that boosting performs better in most cases. (2) Boosting requires less time as compare to RF but more than RUF to decide the best parameter setting. (3) Boosting aims to decrease the bias. Hence, large difference can be noticed between validation and testing accuracy. (4) RUF performs better than RF as it involves less correlated trees. (4) Highest accuracy is achieved with adaboost using class-specific feature set. 

\subsection{Experiment with intermediate images}
The low contrast of intermediate samples characterizes a low disease severity and indicates an early stage sample. Thus, correct identification of intermediate samples is important for early diagnosis. This highlights the clinical importance of such an experiment. Here, intermediate samples which were not a part of training set are taken for testing while training set contains the positive as well as intermediate images. The classification performance for all the classes are
illustrated in Table~\ref{table10}, for some well performing classification frameworks. This demonstrate that the results for the intermediate samples almost match those with all samples. Hence, the proposed approach is very effective even for low-contrast images.

\begin{table}[h!]
	\centering
	\caption{Performance of intermediate samples: Testing (figures in \%).}
	\label{table10}
	\medskip
	\setlength\extrarowheight{1pt} 
	\resizebox{13.80cm}{!}{
		\begin{tabular}{|c|c|c|c|c|c|c|c|c|c|c|}
			\hline
			\multirow{3}{*}{\textbf{}} & \multicolumn{10}{c|}{\textbf{Class-specific features}}   \\ \cline{2-11} 
			
			\multirow{3}{*}{} & \multicolumn{2}{c|}{\multirow{2}{*}{\textbf{One vs. one}}} & \multicolumn{2}{c|}{\multirow{2}{*}{\textbf{Random Forest}}} & \multicolumn{2}{c|}{\multirow{2}{*}{\textbf{Adaboost}}} & \multicolumn{4}{c|}{\textbf{One stage hierarchy}}                                   \\ \cline{8-11} 
			& \multicolumn{2}{c|}{}                            & \multicolumn{2}{c|}{}                            & \multicolumn{2}{c|}{}                            & \multicolumn{2}{c|}{\textbf{First Stage}} &  \multicolumn{2}{c|}{\textbf{\begin{tabular}[c]{@{}c@{}}Second stage \\(using SVM score)\end{tabular}}} \\ \cline{2-11}

			& \textbf{OTP}        & \textbf{OFP}       & \textbf{OTP}      & \textbf{OFP}      & \textbf{OTP}        & \textbf{OFP}       & \textbf{OTP}      & \textbf{OFP}      & \textbf{OTP}         & \textbf{OFP}         \\ \hline
			\textbf{Class1}            & 99.14               & 0.19              & 98.39             & 0.38              & 98.72               & 0.30               & 99.47             & 2.72              & 99.47              & 2.72                            \\ \hline
			\textbf{Class2}            & 95.78               & 0.57              & 95.45             & 1.34               & 97.95              & 0.60               & 95.48             & 0.89              & 95.47               & 0.53                          \\ \hline
			\textbf{Class3}            & 97.70              & 1.21              & 95.08            & 1.41             & 97.50              & 0.72              & 97.42            & 1.14              & 97.37              &0.96                           \\ \hline
			\textbf{Class4}            & 98.44               & 0.19               & 98.54            & 0.33              & 98.68               & 0.30                & 97.90             & 0.22              &97.40               & 0.15                          \\ \hline
			\textbf{Class5}            & 96.03              & 0.75              & 94.41            & 1.16            & 95.94            & 0.85               & 95.29             & 0.79             & 83.17              & 0.55                             \\ \hline
			\textbf{Class6}            & 95.39              & 4.94                & 88.96            & 0.55             & 94.66               & 0.20              & 94.69             & 0.39              & 94.49              & 0.39                            \\ \hline
			\textbf{Avg.}              & 97.08              & 1.30               & 95.14             & 0.86              & 97.24              & 0.51               & 96.70             & 1.02              & 94.56               & 0.89                        \\ \hline
			\textbf{F-score}           & \multicolumn{2}{c|}{\textbf{0.9787}}     & \multicolumn{2}{c|}{\textbf{0.9708}}  & \multicolumn{2}{c|}{\textbf{0.9835}}     & \multicolumn{2}{c|}{\textbf{0.9782}}     & \multicolumn{2}{c|}{\textbf{0.9729}}   \\ \hline
	\end{tabular}}
\end{table}

\subsection{Comparison among all classification frameworks} 
Table~\ref{table8} provides comparable performance among all the classification frameworks (class-specific features) under one roof. It illustrates the comparable performance of all frameworks using class-specific feature set (for testing). In table, for the two-stage methods, we have only included the results obtained at the end of second stage. Form this, it can be observed that (1) Class-specific features that designed thoughtfully produce higher accuracy than texture and combined feature sets in all considered frameworks. (2) From complexity perspective, the two-stage frameworks are more complex than other frameworks. However, the first stage in the hierarchical approach has a lower complexity (in terms of feature space), and in general, the first stage methods can also be treated as differential diagnosis. (3) Common hierarchical with first stage performs better than the other two-stage approaches.
(4) Ensemble frameworks  produces higher accuracy with less false positive. (4) 

\begin{table}[h!]
	\centering
	\caption{Comparison of frameworks: Class-specific features (testing).}
	\label{table8}
	\medskip
	\setlength\extrarowheight{4.5pt} 
	\resizebox{13.80cm}{!}{
		\begin{tabular}{|c|c|c|c|c|c|c|c|}
			\hline
			\multirow{2}{*}{} & \multicolumn{7}{c|}{\textbf{Classification Frameworks: Class-specific features (figures in \%)}}                                                                                                                                                                                                                                               \\ \cline{2-8} 
			& \textbf{\begin{tabular}[c]{@{}c@{}}Two-stage \\ cascade system\end{tabular}} & \textbf{\begin{tabular}[c]{@{}c@{}}Common\\  hierarchy\end{tabular}} & \textbf{One vs. rest} & \textbf{One vs. one} & \textbf{Random forest} & \textbf{\begin{tabular}[c]{@{}c@{}}Random uniform \\ forest\end{tabular}} & \textbf{Adaboost} \\ \hline\hline
			
			& \multicolumn{7}{c|}{Second Stage}   \\ \hline\hline
			\textbf{OTP}      & 94.14                                                                        & 97.11                                                                   & 96.52                & 97.68                & 96.64               &97.42                                                                     & 98.18            \\ \hline
			\textbf{OFP}      & 1.55                                                                        & 0.37                                                                     & 0.77                  & 0.40                 & 0.62                   & 0.46                                                                      & 0.34             \\ \hline\hline

			\textbf{F-score }          & \textbf{.9644}                                                               & \textbf{.9835}                                                               & \textbf{.9786}       & \textbf{.9863}        & \textbf{.9768}              & \textbf{.9840}                                                                 & \textbf{.9891}         \\ \hline
	\end{tabular}}
\end{table}

\subsection{Comparison with contemporary approaches}
We provide a brief discussion placing our work in perspective of the recent methods~\cite{Zhimin}-\cite{Donato} ,~\cite{Manivannan}-\cite{S}  which use the same dataset as us. These approaches employ the standard pipeline with standard sophisticated feature (common) features across classes followed by a classification framework (often SVM). For brevity, we compute and provide only the average results from these approaches in Table~\ref{table9}. Our defined feature set clearly outperforms the state-of the-art methods in most of the classification frameworks. The methods~\cite{rodrigues2017,jia2016} which produces slightly higher accuracy than our utilize CNN framework. However, in terms of complexity, proposed framework is much simpler than CNN. 

Thus, such a comparison demonstrates the effectiveness of class-specific feature set in various classification framework. The proposed work yields encouraging results with respect to the state-of-the-art, and highlights the role of specific features in different classification framework.

\begin{table}[h!]
	\centering
	\caption{Performance comparison (figures in \%).}
	\label{table9}
	\medskip
	\setlength\extrarowheight{6pt} 
	\resizebox{8.44cm}{!}{
		\begin{tabular}{|c|c|c|c|c|}
			\hline
			\multicolumn{2}{|c|}{\textbf{Methods}}            & \textbf{OTP(\%)}                                                & \textbf{OFP(\%)}  \\ \hline
			\multicolumn{2}{|c|}{Zhimin et al.~\cite{Zhimin}}               & 96.76                                                           & 3.24        \\ \hline
			\multicolumn{2}{|c|}{Larissa et al.~\cite{rodrigues2017}}               & 98.17                                                           & 5.69             \\ \hline 
			\multicolumn{2}{|c|}{Xi et al.~\cite{jia2016}}               & 98.26                                                                 & \textbf{-}       \\ \hline
			\multicolumn{2}{|c|}{Nanni et al.~\cite{L}}                & 60.9                                                            & NA               \\ \hline
			\multicolumn{2}{|c|}{Ilias et al.~\cite{T}}                & 89.6                                                            & 9.66           \\ \hline
			\multicolumn{2}{|c|}{Shahab et al.~\cite{Ensafi}}               & 94.9                                                            & 5.09            \\ \hline
			\multicolumn{2}{|c|}{Diego et al.~\cite{Gragnaniello}}                & 81.10                                                           & 18.89           \\ \hline
			\multicolumn{2}{|c|}{Cascio et al.~\cite{Donato}}               & 84.58                                                           & NA               \\ \hline
			
			\multicolumn{2}{|c|}{Siyamalan et al.~\cite{Manivannan}}            & 95.21                                                           & 7.38            \\ \hline
			\multicolumn{2}{|c|}{Siyamalan et al.~\cite{S}}            & 80.25                                                           & 19.37          \\ \hline \hline
			
			\multirow{14}{*}{\begin{tabular}[c]{@{}c@{}c@{}}Proposed \\ approach\end{tabular}} & Two-stage cascade system  & \begin{tabular}[c]{@{}c@{}}CTP-1:39.38\\ ATP-1: 55.41\\ OTP-2: 94.14\end{tabular}   & 1.55           \\ \cline{2-4} 
			& Common hierarchy & \begin{tabular}[c]{@{}c@{}} CTP-1: 96.03\\ ATP-1: 2.42\\OTP-2: 97.11  \end{tabular} & 0.37          \\ \cline{2-4}
			& One vs. rest & \begin{tabular}[c]{@{}c@{}}CTP-1: 90.53 \\ ATP-1: 7.50 \\ OTP-2: 96.52\end{tabular} & 0.77      \\ \cline{2-4}
			& One vs. one & \begin{tabular}[c]{@{}c@{}}OTP:97.68\end{tabular} &0.40           \\ \cline{2-4}
			
			& Random Forest & \begin{tabular}[c]{@{}c@{}} OTP: 96.64\end{tabular} & 0.62        \\ \cline{2-4}
			& Random Uniform Forest (RUF) & \begin{tabular}[c]{@{}c@{}} OTP: 97.42\end{tabular} & 0.46        \\ \cline{2-4}
			& Adaboost & \begin{tabular}[c]{@{}c@{}} OTP: 98.18\end{tabular} &0.34        \\ \hline \hline
	\end{tabular}}
\end{table}

We note that many of these approaches involve more sophisticated feature extraction methods (SIFT, SURF, LBP, HOG etc.). In comparison our feature extraction is quite simple and arguably more efficient, and can still provide a high TP rate, and a reasonably low FP rate. Note that cascade approach can be easily adapted to include more features, while maintaining the hierarchical classification framework in the first stage. However, as we demonstrate, even with the relatively simplistic features that we employ, our approach is able to outperforms better than most state-of-the-art methods. 

\section{Conclusions and future work}
In this paper, we employ  simple, efficient and visually more interpretable,  class-specific features which defined based on the visual characteristics of  each class. Considering that the problem consists of few classes and  the simplicity of proposed features, we consider variants of  classification frameworks which are designed as an ensembles of various binary classifiers. In addition to providing detailed discussions on their (frameworks) silent aspects, and pros and cons over each other, we perform various experiments which include different feature sets and demonstrate the effectiveness of class-specific feature set in various classification frameworks. We make insightful comparisons between different types of classification frameworks. We also demonstrate the validity of considered feature-framework by an experiment which considers the challenging case of intermediate images. The proposed work yields encouraging results with respect to the state-of-the-art and highlights the role of class-specific features in different classification framework. This work provides a new direction for structuring the HEp-2 cell image classification in terms of using simple and visually interpretable class-specific features. Moreover, it also serves as a review and demonstration of a variety of classification strategy for the task of HEp-2 cell image classification. Our future work would involve exploring automated approaches for feature selection for individual blocks.




\bibliography{bare_jrnl}   
\bibliographystyle{spiejour}   

%
%
%
%
%
%

\end{spacing}
\end{document}